\documentclass{article}

\PassOptionsToPackage{numbers, compress}{natbib}

\usepackage[final]{neurips_2024}

\usepackage[utf8]{inputenc} 
\usepackage[T1]{fontenc}    
\usepackage{hyperref}       
\usepackage{url}            
\usepackage{booktabs}       
\usepackage{amsfonts}       
\usepackage{nicefrac}       
\usepackage{microtype}      
\usepackage[table]{xcolor}         

\usepackage{graphicx}
\usepackage{subcaption}
\usepackage{amsmath}
\usepackage{amssymb}
\usepackage{dsfont}
\usepackage{algorithm}
\usepackage{algorithmic}
\usepackage{multirow}
\usepackage{array}
\usepackage{caption}
\usepackage{listings}
\usepackage[frozencache]{minted}
\usepackage{mystyle}
\usepackage{titlesec}
\usepackage{marvosym}

\titlespacing\section{0pt}{12pt plus 4pt minus 2pt}{0pt plus 2pt minus 2pt}

\newcommand{\myparagraph}[1]{\noindent\textbf{#1}}
\definecolor{lightgray}{gray}{0.9}

\title{GraphMorph: Tubular Structure Extraction by Morphing Predicted Graphs}

\author{%
  Zhao Zhang$^{1,5}$ \quad Ziwei Zhao$^{2}$ \quad Dong Wang$^{2}$ \quad \textbf{Liwei Wang}$^{3,4,\textrm{\Letter}}$\\
  $^1$Center for Data Science, Peking University \quad
  $^2$Yizhun Medical AI Co., Ltd \\
  $^3$State Key Laboratory of General Artificial Intelligence,\\ School of Intelligence Science and Technology, Peking University\quad \\
  $^4$Center for Machine Learning Research, Peking University \quad
  $^5$Pazhou Laboratory (Huangpu) \\
  \fontsize{9pt}{0pt}{\texttt{zhangzh@stu.pku.edu.cn} \quad \texttt{ziwei.zhao@yizhun-ai.com}} \\
  \fontsize{9pt}{0pt}{\texttt{dong.wang@yizhun-ai.com} \quad\quad \texttt{wanglw@pku.edu.cn}\quad\quad}
}

\begin{document}

\maketitle

\begin{abstract}
  Accurately restoring topology is both challenging and crucial in tubular structure extraction tasks, such as blood vessel segmentation and road network extraction. Diverging from traditional approaches based on pixel-level classification, our proposed method, named GraphMorph, focuses on branch-level features of tubular structures to achieve more topologically accurate predictions. GraphMorph comprises two main components: a Graph Decoder and a Morph Module. Utilizing multi-scale features extracted from an image patch by the segmentation network, the Graph Decoder facilitates the learning of branch-level features and generates a graph that accurately represents the tubular structure in this patch. The Morph Module processes two primary inputs: the graph and the centerline probability map, provided by the Graph Decoder and the segmentation network, respectively. Employing a novel \(\mathrm{SkeletonDijkstra}\) algorithm, the Morph Module produces a centerline mask that aligns with the predicted graph. Furthermore, we observe that employing centerline masks predicted by GraphMorph significantly reduces false positives in the segmentation task, which is achieved by a simple yet effective post-processing strategy. The efficacy of our method in the centerline extraction and segmentation tasks has been substantiated through experimental evaluations across various datasets. Source code will be released soon.
\end{abstract}

\section{Introduction}
\label{sec:intro}

\looseness=-1 Extraction of tubular structures is an essential step in many computer vision tasks~\cite{staal2004ridge, hoover2000locating, arganda2015crowdsourcing, mnih2013machine}. In medical applications, accurate segmentation of retinal vessels can provide crucial insights into various
cardiovascular and ophthalmologic diseases~\cite{fraz2012blood}. In the field of urban planning and geographic information systems, the precise extraction of road networks aids in traffic management, urban development, and emergency response planning~\cite{mnih2010learning}. Existing deep learning-based methods model tubular structure extraction as a pixel-level classification task~\cite{ronneberger2015u} or point set prediction task~\cite{wang2022pointscatter}, without explicitly predicting the topological structures. To focus more on topology, some advanced methods design novel backbones or modules~\cite{shin2019deep, mei2021coanet, qi2023dynamic}, or introduce new loss functions from the topological perspective~\cite{hu2019topology, shit2021cldice, menten2023skeletonization}. However, they are still limited to the framework of pixel-level prediction.

\begin{figure}
  \centering
  \includegraphics[width=0.9\textwidth]{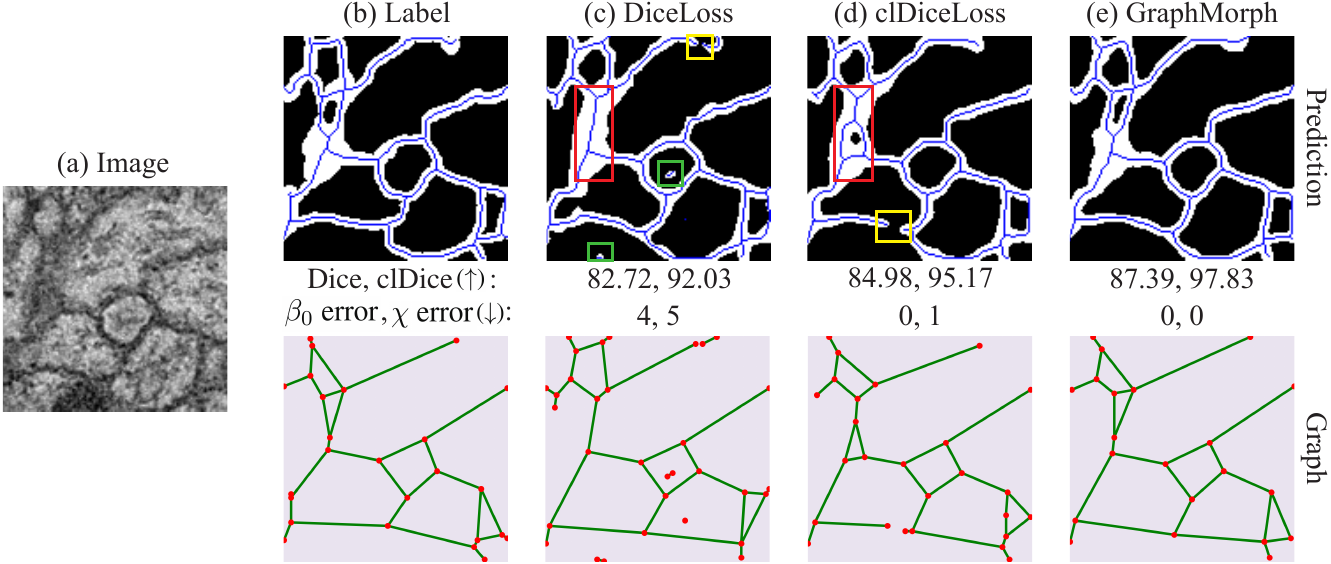}
  \caption{Illustrating the impact of topological feature utilization on segmentation accuracy.  \textbf{(a)} An input neuron image. \textbf{Column (b)} Ground truth with segmented membranes (white) and its centerline (blue lines); the constructed graph (nodes in red, edges in green). \textbf{Column (c)} and \textbf{(d)} Predictions of two methods~\cite{milletari2016v, shit2021cldice} without explicit topological learning, highlighting broken branches (false negatives in yellow), redundant branches (false positives in green), and topological errors (in red). \textbf{Column (e)} Our GraphMorph guarantees topological accuracy by learning explicit branch-level features. Details of skeletonization and graph construction are given in Appendix~\ref{ssec:graph_construction}. Evaluation metrics: Dice and clDice (higher is better), \(\beta_0\) error and \(\chi\) error (lower is better).}
  \label{fig:intro_compare}
\vspace{-10pt}
\end{figure}

We argue that most pixel-level frameworks have not effectively exploited the nature of tubular structures, which are inherently composed of several branches that are interconnected in complex ways. Specifically, pixel-level loss functions, like softDice Loss~\cite{milletari2016v} and Focal Loss~\cite{lin2017focal}, struggle with subtle inaccuracies and are particularly ineffective at addressing complex topological errors. Under the pixel-level frameworks, despite attempts to pay more attention to fine branches~\cite{shit2021cldice, menten2023skeletonization} or topological features~\cite{hu2019topology,qi2023dynamic}, they still struggle with fully capturing the complex topological nature of tubular structures. We demonstrate this deficiency by providing an example in Figure~\ref{fig:intro_compare}. For a systematic understanding of the issues of pixel-level frameworks, we summarize them into three categories: (1) Broken branches or false negatives (FNs). (2) Redundant branches or false positives (FPs). (3) Topological errors (TEs). Therefore, understanding tubular structure extraction solely from a pixel-level perspective is fundamentally flawed.

Recognizing these limitations, we shift our focus to \textbf{branch-level features}, which are more essential for accurately capturing the nuances of tubular structures. Any complex tubular structure can be broken down into several branches, which distinguishes it from non-tubular objects. This perspective inspires us to extract tubular structures in two steps: (1) predicting the location of the two endpoints of each branch; (2) finding the optimal path between the two endpoints of each branch. Such a solution is intuitively aligned with human perception, and able to offer several advantages. Specifically, if the endpoints of branches are accurately predicted in the first step, redundant branches (FPs) are then potentially reduced; and the second step ensures that there is a path connecting the two endpoints of each branch, so that broken branches (FNs) and TEs are effectively suppressed. Besides, learning branch-level features during training elevates the model's focus on topology, which implicitly improves topological accuracy.

To effectively utilize branch-level features of tubular structures, we propose GraphMorph, a pipeline for obtaining topologically accurate centerline masks. GraphMorph consists of a Graph Decoder and a training-free Morph Module, which corresponds to the two steps of our solution respectively. The Graph Decoder, given multi-scale features extracted from an image patch by a segmentation network, predicts the graph \(G\) of the tubular structure. \(G\) is defined by a node set \(V=\{(x_i, y_i)\}_{i=1}^N\), which contains the coordinates of all critical points vital for maintaining topology, and an adjacency matrix \(A \in \{0, 1\}^{N \times N}\), which encodes the connectivity among nodes. Each pair of connected nodes in the graph corresponds to two endpoints of a branch of the tubular structure, thus the Graph Decoder takes full advantage of branch-level features through this graph representation. Technically, the prediction of \(V\) is addressed as a set prediction problem, solvable by our modified version of Deformable DETR~\cite{zhu2020deformable}. To efficiently obtain the adjacency matrix \(A\), we design a lightweight link prediction module that capitalizes on the extracted node features. Concretely, since the number of nodes in each tubular structure may be different, we generate linear weights and biases dynamically conditioned on node features, and the adjacency list of each node is obtained from its corresponding linear parameters (See Figure~\ref{fig:methodology}).

\looseness=-1 The Morph Module, a core contribution of this work, is intended to obtain topologically accurate centerline masks. While studies in the image-to-graph task~\cite{shit2022relationformer, prabhakar2024vesselformer} also utilize graph representation, they struggle to directly obtain accurate centerline masks due to the curved nature of tubular objects. In contrast, our Morph Module generates topologically accurate centerline masks via a novel \(\mathrm{SkeletonDijkstra}\) algorithm. Specifically, a centerline probability map \(P_m\), together with the graph \(G\), output by the segmentation network and the Graph Decoder respectively, serve as the input to the Morph Module. Afterwards, considering the skeleton property of centerlines, our \(\mathrm{SkeletonDijkstra}\) algorithm finds the optimal path between each pair of connected nodes. In particular, during path finding from the start point to the end point, we always restrict the path to a single pixel width to satisfy the skeleton property of centerlines. Consequently, the topology of the resulting centerline mask is guaranteed by \(G\), leading to a reduction in TEs. This method also minimizes the occurrence of broken branches (FNs) and redundant branches (FPs), which is a significant improvement over direct pixel-level operations on \(P_m\), such as thresholding.

We conduct the experiments by beginning with the \textbf{centerline extraction} task to verify the effects of the two components of GraphMorph. Experimentally, serveing as an auxiliary training module to learn the graph representation, the Graph Decoder enhances the segmentation network's focus on branch-level features, thus both volumetric metrics
and topological metrics are boosted. Furthermore, employing the Morph Module at inference stage considerably improves topological metrics. For the \textbf{segmentation} task, we develop a streamlined post-processing strategy to refine segmentation masks via the topologically accurate centerline masks output by the Morph Module, significantly suppressing false positives of segmentation results. To verify the effectiveness of GraphMorph and the post-processing strategy, we conduct extensive experiments across four typical tubular structure extraction datasets. We have applied our methodology on three powerful backbones and achieved consistent improvements in all metrics. Moreover, compared with the state-of-the-art methods, our approach achieves the best results across all datasets.

In a nutshell, our contributions can be summarized as the following: (1) We introduce GraphMorph, an innovative framework specifically tailored for tubular structure extraction. Based on the proposed Graph Decoder and Morph Module, the branch-level features are fully exploited and the topologically accurate centerline masks are derived naturally. (2) For the segmentation task, an efficient post-processing strategy significantly suppresses false positives via the centerline masks predicted by GraphMorph, ensuring that the segmentation results are more closely aligned with the predicted graphs. (3) Experimental results on three medical datasets and one road dataset underscore the effectiveness of our method. For both centerline extraction and segmentation tasks, GraphMorph has achieved remarkable improvements across all metrics, especially in topological metrics.

\section{Related Work}

\myparagraph{Image segmentation of tubular structures.}
Deep learning-based methods have achieved impressive results in segmentation tasks~\cite{long2015fully, ronneberger2015u, chen2018encoder}. To further enhance the segmentation performance of tubular structures, novel network architectures~\cite{jin2019dunet, lou2021dc, shin2019deep, wang2020deep, mei2021coanet, yang2023directional, wang2022pointscatter, qi2023dynamic} and topology-preserving loss functions~\cite{hu2019topology, mosinska2018beyond, shit2021cldice, menten2023skeletonization, qi2023dynamic} have been proposed. For example, in terms of network architecture, DSCNet~\cite{qi2023dynamic} utilizes dynamic snake convolution to capture fine and tortuous local features; PointScatter~\cite{wang2022pointscatter} explores the point set representation of tubular structures and introduces a novel greedy-based region-wise bipartite matching algorithm to improve training efficiency. In terms of loss functions, clDice~\cite{shit2021cldice} proposes a differentiable soft skeletonization method and achieves loss calculation at centerline level, which implicitly helps model focus more on the fine branches; TopoLoss~\cite{hu2019topology} and TCLoss~\cite{qi2023dynamic} measure the topological similarity of the ground truth and the prediction via persistent homology. Despite these advancements, all of the above methods are still confined to the framework of pixel-level classification and can not entirely overcome their inherent limitations. Our method attempts to morph the predicted graphs of tubular structures to let the network focus more on branch-level features, thus ensuring the topological accuracy of predictions.


\myparagraph{Image to graph.} There are two mainstream subtasks in this area: road network graph detection~\cite{he2020sat2graph, xu2022rngdet, shit2022relationformer, xu2023rngdet++, hetang2024segment} and scene graph generation~\cite{khandelwal2022iterative, kundu2023ggt}. These tasks usually entail detecting key components as nodes (i.e., key points in roads, objects in scenes) and determining their interrelations as edges (i.e., connectivity in roads, interactions in scenes).
Our work differs from these approaches in three ways. 
Firstly, we use only junctions and endpoints as nodes, which allows for explicit semantic characterization of nodes in our graph representation, unlike road network detection tasks where path points may also be regarded as nodes. Secondly, considering the curved nature of tubular objects, we propose Morph Module to obtain topologically accurate centerline masks, a goal that is not addressed by these works. Finally, our dynamic link prediction module is time-efficient, compared with elaborate and time-consuming designs in these works, such as [\texttt{rln}]-token in RelationFormer~\cite{shit2022relationformer}. For a clear understanding, we experimentally compare the differences between our approach and RelationFormer in Appendix~\ref{appendix_link_prediction}. These distinctions make our model not only time-efficient but also applicable to the task of tubular structure extraction with more complex topology.

\section{Method}
\label{sec:Method}

\looseness=-1 This section provides a detailed description of the training and inference procedures of GraphMorph. Figure \ref{fig:methodology} illustrates the training process of our approach, where the segmentation network and Graph Decoder are included. We detail these two components in Section~\ref{ssec:segmentation_network} and \ref{ssec:network_architecture}, respectively. The training details are given in Section \ref{ssec:training_details}. Section \ref{ssec:inference_process} introduces the algorithmic flow of the Morph Module, followed by the inference processes for the centerline extraction and segmentation tasks.

\subsection{Segmentation Network}
\label{ssec:segmentation_network}
The segmentation network processes an input image \(I\) with shape \(\mathbb{H} \times \mathbb{W}\). It serves two purposes: (1) outputting a probability map of tubular structures; (2) providing multi-scale features for the Graph Decoder. For training efficiency, we randomly sample \(R\) regions of interest (ROIs) with size \(H \times H\) in the feature maps (\(R = 3\) in Figure \ref{fig:methodology} for illustration). The ROI is defined as any region containing centerline points. The adoption of ROIs brings two key benefits: it reduces the model's learning complexity due to simpler topological structures within each ROI, and improves training efficiency by decreasing the number of feature tokens processed in the transformer. Technically, We adopt ROI Align~\cite{he2017mask} to extract multi-scale ROI features. Note that the generality of GraphMorph allows it to be adapted to any type of segmentation network. In the experimental part, we validate the enhancement of GraphMorph on a variety of segmentation networks.

\subsection{Graph Decoder}
\label{ssec:network_architecture}

\begin{figure}
  \centering
  \includegraphics[width=\textwidth]{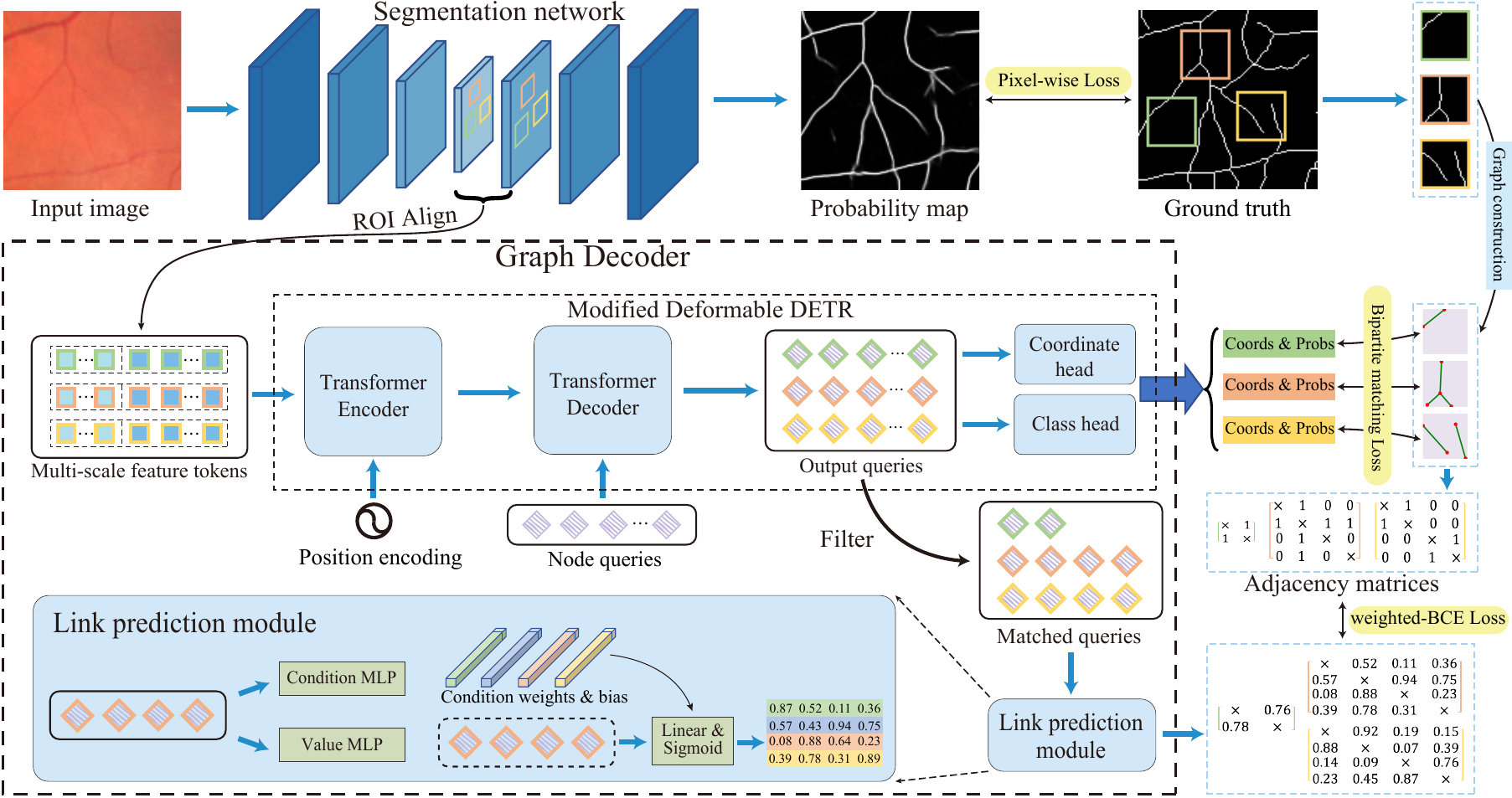}
  \caption{Overview of the training process. Given an image, the segmentation network outputs a probability map of the centerline or segmentation and produces multi-scale feature maps. Then, \(R\) regions of interest (ROIs) are randomly sampled from the image, and their corresponding features are fed into the Graph Decoder, which predicts the nodes within these ROIs using a modified Deformable DETR and outputs the adjacency matrices utilizing the proposed link prediction module.}
  \label{fig:methodology}
\vspace{-10pt}
\end{figure}

The Graph Decoder is intended to predict the graph for each ROI. Specifically, the modified Deformable DETR~\cite{zhu2020deformable} is responsible for detecting the nodes, while the link prediction module handles predicting the connectivity among these nodes. In the following, we will dissect each component to elucidate their roles.

\myparagraph{Modified Deformable DETR.} We have made two adjustments to the standard Deformable DETR~\cite{zhu2020deformable}. Firstly, since the targets are nodes with only 2-dimensional coordinates, we replace the original box head with a coordinate head that outputs 2-dimensional vectors. Secondly, considering the typically small size of ROIs, we reduce the number of layers in the transformer encoder to three while keep the decoder at six layers. Note that each ROI is treated as an independent sample and different ROIs will not interact with each other in the whole training process. Formally, the process of node prediction can be expressed as follows:
\vspace{-5pt}
\begin{gather}
    \widehat{F}^r=\mathrm{TransformerEncoder}\left(F^r,PE\right) \\
    \widehat{Q}^r=\mathrm{TransformerDecoder}\left(\widehat{F}^r,Q\right) \\
    \hat{s}^r=\mathrm{Sigmoid}\left(\mathrm{ClassHead}\left(\widehat{Q}^r\right)\right), \quad \hat{v}^r=\mathrm{Sigmoid}\left(\mathrm{CoordHead}\left(\widehat{Q}^r\right)\right)
\end{gather}
where \(r=1,2,...,R\). \(F^r \in \mathbb{R}^{L \times C}\) denotes the multi-scale features of the \(r\)-th ROI, and \(Q \in \mathbb{R}^{K \times C}\) is the initial node queries, where $L$ and $K$ are the scale of feature maps and the number of node queries respectively. 
\(PE\) is the multi-scale sinusoidal positional encoding used in \cite{zhu2020deformable}. \(\mathrm{ClassHead}\) is a single linear layer, and \(\mathrm{CoordHead}\) is a 3-layer multilayer perceptron (MLP). \(\hat{s}^r \in \mathbb{R}^{K}\) and \(\hat{v}^r \in \mathbb{R}^{K \times 2}\) are the classification scores and coordinates of the nodes for the \(r\)-th ROI respectively.

\myparagraph{Link prediction module.} Since the number of nodes for each ROI may be different, we design a dynamic module generating linear weights and biases conditioned on node features to directly predict the adjacency matrix \(A\). For the \(r\)-th ROI sample, \(\widehat{Q}^r \in \mathbb{R}^{K \times C}\) is the output queries of the Transformer decoder. The queries matched with the ground truth nodes (the number is denoted as \(P_r\)) during the bipartite matching process are preserved, and the rest queries are filtered out. We denote the kept queries as \(\widetilde{Q}^r \in \mathbb{R}^{P_r \times C}\). As depicted in Figure~\ref{fig:methodology}, the matched queries \(\widetilde{Q}^r\) will be fed into two MLPs. The \(\mathrm{Condition MLP}\) generates a (\(C+1\))-dimensional vector for each matched query, which serves as the weights and biases of the condition linear layer. The \(\mathrm{Value MLP}\) maps the queries to a value space. With the values as input, the condition linear layer of the \(p\)-th query generates the adjacency list of it. The process can be formulated as follows:
\begin{gather}
    W^r_p=\mathrm{ConditionMLP}(\widetilde{Q}^r_p) \in \mathbb{R}^{C+1}, \quad V^r=\mathrm{ValueMLP}(\widetilde{Q}^r) \in \mathbb{R}^{P_r \times C} \label{eq:link1} \\
    \widetilde{A}^r_p=\mathrm{Sigmoid}([W^r_p]_{1:C} \cdot V^r + [W^r_p]_{C+1}) \in\mathbb{R}^{P_r} \label{eq:link3} \\
    \widetilde{A}^r=[(\widetilde{A}^r_1)^T, (\widetilde{A}^r_2)^T, ..., (\widetilde{A}^r_{P_r})^T]^T  \in\mathbb{R}^{P_r \times P_r}  \label{eq:link4}
\end{gather}
where \(p = 1, 2, ..., P_r\). Here, \(\widetilde{Q}^r_p\) denotes the \(p\)-th item of \(\widetilde{Q}^r\), and \(C\) is its dimension. \(W^r_p\) refers to the linear parameters conditioned on the \(p\)-th matched query. \(\widetilde{A}^r_p\) represents the predicted adjacency list for the \(p\)-th matched query, and the concatenation of all lists forms the final adjacency matrix \(\widetilde{A}^r\).

\subsection{Training Details}
\label{ssec:training_details}
\myparagraph{Graph construction.} To train the Graph Decoder, we represent the ground truth of each ROI as a graph (see Figure~\ref{fig:methodology}). The detailed graph construction process can be found in Appendix~\ref{ssec:graph_construction}.

\myparagraph{Label assignment based on bipartite matching.} Bipartite matching is widely used in solving set prediction problems~\cite{carion2020end, zhu2020deformable, wang2022pointscatter}. As in \cite{zhu2020deformable}, we first calculate the cost between the predicted and ground truth nodes. The predicted nodes are denoted as \(\hat{y}=\{(\hat{s}_k,\hat{v}_k)\}_{k=1}^K\), where we omit the index \(r\) of the ROI sample for simplicity. Under the general assumption that \(K\) is larger than the  number of ground truth nodes \(P_r\), thus we pad the set of ground truth nodes with \(\varnothing\) (no node) to achieve a size of \(K\). The ground truth set can be denoted as \({y=\{(c_i, v_i)\}_{i=1}^K}\), where \(c_i\) is the target class label and \(v_i \in [0,1]^2\) is the coordinate of the node. For a permutation \(\sigma \in \mathfrak{S}_K\), where \(\hat{y}_{\sigma(i)}\) is assigned to \(y_i\) (\(i=1,2,..,K\)), we define the cost between \(y_i\) and \(\hat{y}_{\sigma(i)}\) as:
\begin{equation}
    \mathcal{L}_{\text{match}}(y_i, \hat{y}_{\sigma(i)})
    =\lambda_{\text{class}} \cdot \mathds{1}_{\{c_i \neq \varnothing\}}\mathcal{L}_{\text {class}}(\hat{s}_{\sigma(i)})
    +\lambda_{\text{coord}} \cdot\mathds{1}_{\{c_i \neq \varnothing\}} \mathcal{L}_{\text {coord}}(v_i, \hat{v}_{\sigma(i)})
\end{equation}
where \(\lambda_{\text{class}}\) and \(\lambda_{\text{coord}}\) are hyperparameters.
\(
    \mathcal{L}_{\text {class}}(\hat{s}_{\sigma(i)})=\mathcal{L}_{\text {focal}}(\hat{s}_{\sigma(i)}, 1) - \mathcal{L}_{\text {focal}}(\hat{s}_{\sigma(i)}, 0)
\), \(\mathcal{L}_{\text {focal}}(s,c)\) is defined as \(-\alpha \cdot (1-s)^\gamma\log(s)\) if \(c=1\), and \(-(1-\alpha) \cdot s^\gamma\log(1-s)\) if \(c=0\), where \(\alpha\) and \(\gamma\) are hyperparameters.
\(\mathcal{L}_{\text {coord}}\) is commonly used \(\ell_1\) loss. The optimal \(\hat{\sigma}\) is defined as
\vspace{-5pt}
\begin{equation}
    \hat{\sigma}=\underset{\sigma \in \mathfrak{S}_K}{\arg \min } \sum_{i=1}^K \mathcal{L}_{\operatorname{match}}\left(y_i, \hat{y}_{\sigma(i)}\right)
    \vspace{-10pt}
\end{equation}

This optimal assignment can be efficiently obtained by Hungarian algorithm~\cite{kuhn1955hungarian}.

\myparagraph{Loss functions.} To train the Graph Decoder, the overall loss function is comprised of three components: pixel-wise loss \(\mathcal{L}_{\text {Pixel}}\) between the probability map and the ground truth binary mask (in this work, we use softDice~\cite{milletari2016v} and clDice~\cite{shit2021cldice}), Hungarian bipartite matching loss \(\mathcal{L}_{\text {Hungarian}}\) between the predicted and ground truth nodes, weighted-BCE loss \(\mathcal{L}_{\text {Adjacency}}\) between the predicted and ground truth adjacency matrices of the matched queries. For an image with \(R\) ROI samples, the last two loss functions are defined as:
\begin{equation}
    \mathcal{L}_{\text {Hungarian}}(y, \hat{y})=\sum_{r=1}^{R}\sum_{i=1}^{K}\left[\lambda_{\text{class}} \cdot \mathcal{L}_{\text {focal}}(\hat{s}_{\hat{\sigma}(i)}^r, c_i^r)+\lambda_{\text{coord}}\cdot\mathds{1}_{\left\{c_i^r \neq \varnothing\right\}} \mathcal{L}_{\text {coord}}\left(\hat{v}_{\hat{\sigma}(i)}^r,v_i^r\right)\right],
\end{equation}
\begin{equation}
    \mathcal{L}_{\text {Adjacency}}(y, \hat{y})=\sum_{r=1}^{R}\{\frac{0.5}{N_{\text{pos}}}\sum_{i \neq j}^{P_r}\sum_{j=1}^{P_r}(A^r_{ij}\log\widetilde{A}^r_{ij})+\frac{0.5}{N_{\text{neg}}}\sum_{i \neq j}^{P_r}\sum_{j=1}^{P_r}[(1-A^r_{ij})\log(1-\widetilde{A}^r_{ij})]\},
\end{equation}
where \(N_{\text{pos}}\) is the total number of positive locations in ground truth \(\{A^r\}_{r=1}^R\), and \(N_{\text{neg}}\) is the total number of negative locations. Thus, the overall loss function is \(\mathcal{L}_{\text {total}}=\mathcal{L}_{\text {Pixel}}+\mathcal{L}_{\text {Hungarian}}+\mathcal{L}_{\text {Adjacency}}\).

\subsection{Morph Module and Inference}
\label{ssec:inference_process}
The Morph Module is used to get topologically accurate centerline masks, by morphing the predicted graphs from the Graph Decoder. In this subsection, we first introduce the Morph Module, followed by the inference processes for the centerline extraction and segmentation tasks.

\myparagraph{Morph Module.} We present the algorithmic flow in Algorithm \ref{algo:morph}. In particular, \(G=\{V,E\}\) is the graph of an image patch (same size as an ROI sample), and \(P_m\) is the probability map of centerlines obtained from the segmentation network. We iterate over each edge and use our proposed \(\mathrm{SkeletonDijkstra}\) algorithm to find the optimal path with minimum cost. The union of these paths forms the final centerline mask. 

\(\mathrm{SkeletonDijkstra}\) is modified from Dijkstra algorithm~\cite{dijkstra1959note}. We have made two key adaptations for centerline extraction: (1) To restrict the path to a single pixel width, ensuring the property of the skeleton, we mandate that all path points, except for the start and end points, satisfy \(N=2\) (where \(N\) is the number of centerline points in its eight neighbours, see Appendix~\ref{ssec:graph_construction}). (2) To suppress potential false-positive edges from the Graph Decoder, we exclude the paths with an average cost exceeding a threshold \(p_{thresh}\). These refinements optimize the algorithm to yield topologically accurate centerline masks. The detailed algorithmic flow of \(\mathrm{SkeletonDijkstra}\) can be seen in Algorithm \ref{algo:skeleton_dijkstra} in Appendix~\ref{appendix_SkeletonDijkstra}.

\begin{algorithm}
\caption{Morph Module}
\label{algo:morph}
\begin{algorithmic}
\REQUIRE Node set \( V \), Edge set \( E \), Probability map \( P_m \)
\ENSURE Centerline mask \( M \)

\STATE Initialize \( M \) as a zero matrix with the same size as \( P_m \)
\STATE Initialize \( C_m \) where \( C_m[i][j] = 1 - P_m[i][j] \) for each element
\FORALL{edges \( (u, v) \) in \( E \)}
    \STATE \( path \) $\leftarrow$ \textbf{SkeletonDijkstra}(\( u \), \( v \), \( C_m \), \(p_{thresh}\))
    \FORALL{points \( p \) in \( path \)}
        \STATE Set \( M[p.x][p.y] = 1 \)
    \ENDFOR
\ENDFOR
\RETURN \( M \)
\end{algorithmic}
\end{algorithm}

\begin{figure}
  \centering
  \includegraphics[width=0.9\textwidth]{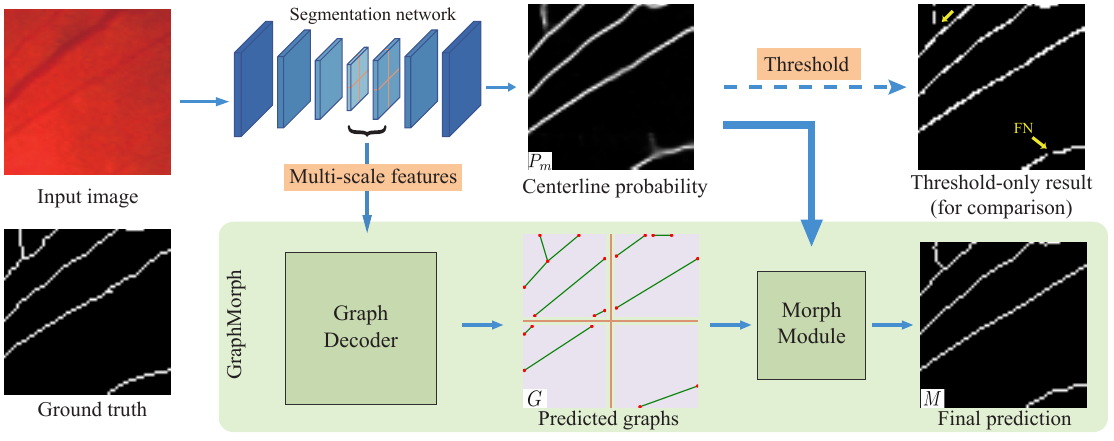}
  \caption{Inference process of centerline extraction. First, the segmentation network generates a centerline probability map \(P_m\) along with multi-scale image features. Subsequently, the Graph Decoder utilizes the image features to predict graphs \(G\) via sliding window inference. Finally, the Morph Module employs \(P_m\) to find the optimal path between each pair of connected nodes in \(G\), resulting in a final centerline mask. This approach achieves higher topological accuracy compared to direct thresholding of \(P_m\).}
  \label{fig:inference_centerline}
\vspace{-10pt}
\end{figure}

\myparagraph{Inference of centerline extraction.} As depicted in Figure~\ref{fig:inference_centerline}, the centerline extraction process begins with generating a centerline probability map via the segmentation network. Then, sliding window inference is employed across the entire image in Graph Decoder to obtain graphs for all split patches.  Finally, the Morph Module produces the centerline mask for each patch, and the combination of these masks forms the complete centerline mask of the entire image.

\myparagraph{Inference of segmentation.} Since the segmentation probability map \(S_m\) can not be used directly by the Morph Module, we first threshold \(S_m\) to obtain segmentation mask \(S_m^\prime\) and skeletonize it into a centerline mask \(P_m^\prime\). The distance from each pixel to the nearest centerline point in \(P_m^\prime\) is calculated and normalized to create a distance map \(D\). Then the centerline probability map \(P_m\) is obatined by \(P_m = S_m \times (1 - D)\). Employing the Morph Module on \(P_m\) yields a topologically precise mask \(M\). To suppress false positives in \(S_m^\prime\) (especially isolated regions), a post-processing strategy is initiated from \(M_0 = M \odot S_m^\prime\). This strategy involves iteratively expanding \( M_0 \) within the boundaries of \( S_m^\prime \) until stabilization. The stabilized mask \( M_T \) is then taken as the final output. This approach, as confirmed by experiments, effectively diminishes false positives and enhances topological accuracy. The above-mentioned soft skeletonization method (from \(S_m\) to \(P_m\)) and post-processing strategy introduce minimal time cost and are straightforward to implement, with detailes in Appendix~\ref{appendix_SoftSke} and Appendix~\ref{appendix_SuppressFPs}.

\section{Experiments}
\label{sec:Experiments}

\subsection{Experimental Setup}
\label{ssec:experimental_setup}
\myparagraph{Datasets.} We evaluate GraphMorph on three medical datasets and one road dataset. DRIVE~\cite{staal2004ridge} and STARE~\cite{hoover2000locating} are retinal vessel datasets commonly used in medical image segmentation. ISBI12~\cite{arganda2015crowdsourcing} contains 30 Electron
Microscopy images to segment membranes. The Massachusetts Roads (MassRoad) dataset contains 1171 aerial images for road network extraction. We use the data splits for DRIVE and STARE provided in MMSegmentation~\cite{mmseg2020}. For ISBI12, following previous works~\cite{seyedhosseini2013image,mosinska2018beyond}, we split it into 15 images for training and 15 for testing. For MassRoad, we follow ~\cite{wang2022pointscatter} to construct the training set, and the total 63 images of the official validation set and test set are used for testing.

\myparagraph{Baselines.} We adopt affluent baselines for comparison, including UNet~\cite{ronneberger2015u}, ResUNet~\cite{zhang2018road}, CS-Net~\cite{mou2019cs}, DC-UNet~\cite{lou2021dc}, TransUNet~\cite{chen2021transunet}, DSCNet~\cite{qi2023dynamic} and PointScatter~\cite{wang2022pointscatter}. Particularly, we use LinkNet34~\cite{chaurasia2017linknet} and D-Linknet34~\cite{zhou2018d} as baselines for the MassRoad dataset. In addition, we compare with TopoLoss~\cite{hu2019topology}, which is a topology-based loss function.

\myparagraph{Metrics.} For the centerline extraction task, we use Dice~\cite{zou2004statistical}, Accuracy (ACC) and AUC as volumetric metrics. For robust evaluation, we give a tolerance of a 5-pixel region around the ground truth centerline mask following~\cite{guimaraes2016fast}. We compute topological metrics following~\cite{wang2022pointscatter}, including the mean absolute errors of \(\beta_0\), \(\beta_1\) and the Euler characteristic. To compare fairly, we skeletonize the prediction before evaluation. For the segmentation task, we adopt Dice, clDice~\cite{shit2021cldice} and ACC as volumetric metrics and the same topological metrics as the centerline extraction task. Moreover, ARI (Adjusted Rand Index)~\cite{hubert1985comparing} and VOI (Variation of Information)~\cite{meilua2007comparing} are used to evaluate clustering similarity.

\myparagraph{Implementation Details.} For three medical datasets, we use randomly cropped \(384 \times 384\) images for training. The size of ROI samples \(H\) is \(32\) and the stride of sliding window used in inference process is \(30\). For MassRoad dataset, the cropped size is \(768 \times 768\), \(H=48\) and the stride is 45. For all experiments, we use 64 ROI samples per image (\(R=64\)) to train the Graph Decoder, and the number of node queries in the modified Deformable DETR is set to 100 (\(K=100\)). According to previous experiences~\cite{shit2022relationformer}, the default hyperparameters used in loss functions are as follows: \(\lambda_{\text{class}}=0.2\), \(\lambda_{\text{coord}}=0.5\), \(\alpha=0.6\), \(\gamma=2\). We use \(\alpha=0.75\) for MassRoad due to the sparse nature of the road networks. For all types of segmentation networks, we use multi-scale features ranging from the lowest resolution to the \(4\times\) downsampling of the original image as input of the Graph Decoder. More implementation details are introduced in Appendix~\ref{appendix_implementation}.

\begin{table}[htbp]
  \caption{Centerline extraction performance on four public datasets based on UNet.}
  \label{tab:cline_results}
  \centering
  \resizebox{\textwidth}{!}{%
  \begin{tabular}{cccccccc}
    \toprule
    \multirow{2}{*}{Dataset} & \multirow{2}{*}{Method} & \multicolumn{3}{c}{Volumetric metrics (\(\uparrow\))} & \multicolumn{3}{c}{Topological metrics (\(\downarrow\))} \\
    \cmidrule(lr){3-5} \cmidrule(lr){6-8}
            &      & Dice & AUC & ACC & \(\beta_0\) error & \(\beta_1\) error & \(\chi\) error \\
    \midrule
    \multirow{4}{*}{DRIVE} 
    & softDice~\cite{milletari2016v} & 0.7353 ± 0.0127 & 0.9333 ± 0.0089 & 0.9768 ± 0.0013 & 2.169 ± 0.112 & 1.590 ± 0.107 & 2.537 ± 0.139 \\
    & PointScatter~\cite{wang2022pointscatter} & 0.7381 ± 0.0133 & 0.9401 ± 0.0078 & 0.9775 ± 0.0013 & 3.259 ± 0.153 & 2.080 ± 0.120 & 3.500 ± 0.176 \\
    & \cellcolor{lightgray}{softDice~\cite{milletari2016v} + Graph Decoder} & \cellcolor{lightgray}{\textbf{0.7506 ± 0.0127}} & \cellcolor{lightgray}{\textbf{0.9481 ± 0.0082}} & \cellcolor{lightgray}{\textbf{0.9783 ± 0.0012}} & \cellcolor{lightgray}{1.552 ± 0.094} & \cellcolor{lightgray}{1.382 ± 0.106} & \cellcolor{lightgray}{1.899 ± 0.125} \\
    & \cellcolor{lightgray}{softDice~\cite{milletari2016v} + Graph Decoder + Morph Module} & \cellcolor{lightgray}{0.7496 ± 0.0118} & \cellcolor{lightgray}{/} & \cellcolor{lightgray}{0.9776 ± 0.0012} & \cellcolor{lightgray}{\textbf{0.555 ± 0.038}} & \cellcolor{lightgray}{\textbf{1.074 ± 0.073}} & \cellcolor{lightgray}{\textbf{0.893 ± 0.061}} \\
    \midrule
    \multirow{4}{*}{ISBI12} 
    & softDice~\cite{milletari2016v} & 0.6428 ± 0.0104 & 0.8937 ± 0.0063 & 0.9737 ± 0.0013 & 4.045 ± 0.191 & 2.696 ± 0.112 & 4.294 ± 0.205 \\
    & Pointscatter~\cite{wang2022pointscatter} & 0.6546 ± 0.0089 & 0.9104 ± 0.0057 & \textbf{0.9747 ± 0.0013} & 6.398 ± 0.277 & 3.156 ± 0.124 & 6.548 ± 0.290 \\
    & \cellcolor{lightgray}{softDice~\cite{milletari2016v} + Graph Decoder} & \cellcolor{lightgray}{0.6486 ± 0.0095} & \cellcolor{lightgray}{\textbf{0.9240 ± 0.0061}} & \cellcolor{lightgray}{0.9742 ± 0.0012} & \cellcolor{lightgray}{4.013 ± 0.179} & \cellcolor{lightgray}{2.732 ± 0.110} & \cellcolor{lightgray}{4.249 ± 0.193} \\
    & \cellcolor{lightgray}{softDice~\cite{milletari2016v} + Graph Decoder + Morph Module} & \cellcolor{lightgray}{\textbf{0.6687 ± 0.0092}} & \cellcolor{lightgray}{/} & \cellcolor{lightgray}{0.9742 ± 0.0014} & \cellcolor{lightgray}{\textbf{0.665 ± 0.049}} & \cellcolor{lightgray}{\textbf{1.207 ± 0.070}} & \cellcolor{lightgray}{\textbf{0.858 ± 0.059}} \\
    \midrule
    \multirow{4}{*}{STARE} 
    & softDice~\cite{milletari2016v} & 0.7119 ± 0.0392 & 0.9290 ± 0.0283 & 0.9889 ± 0.0012 & 1.874 ± 0.139 & 1.209 ± 0.112 & 2.063 ± 0.162 \\
    & Pointscatter~\cite{wang2022pointscatter} & 0.7224 ± 0.0414 & 0.9494 ± 0.0179 & 0.9896 ± 0.0012 & 2.080 ± 0.149 & 1.365 ± 0.116 & 2.213 ± 0.166 \\
    & \cellcolor{lightgray}{softDice~\cite{milletari2016v} + Graph Decoder} & \cellcolor{lightgray}{\textbf{0.7298 ± 0.0428}} & \cellcolor{lightgray}{\textbf{0.9506 ± 0.0208}} & \cellcolor{lightgray}{\textbf{0.9898 ± 0.0011}} & \cellcolor{lightgray}{1.467 ± 0.113} & \cellcolor{lightgray}{1.074 ± 0.104} & \cellcolor{lightgray}{1.654 ± 0.132} \\
    & \cellcolor{lightgray}{softDice~\cite{milletari2016v} + Graph Decoder + Morph Module} & \cellcolor{lightgray}{0.7291 ± 0.0387} & \cellcolor{lightgray}{/} & \cellcolor{lightgray}{0.9894 ± 0.0011} & \cellcolor{lightgray}{\textbf{0.482 ± 0.042}} & \cellcolor{lightgray}{\textbf{0.799 ± 0.077}} & \cellcolor{lightgray}{\textbf{0.653 ± 0.059}} \\
    \midrule
    \multirow{4}{*}{MassRoad} 
    & softDice~\cite{milletari2016v} & 0.6339 ± 0.0169 & 0.9718 ± 0.0047 & \textbf{0.9942 ± 0.0009} & 1.672 ± 0.056 & 1.627 ± 0.087 & 1.968 ± 0.097 \\
    & Pointscatter~\cite{wang2022pointscatter} & \textbf{0.6405 ± 0.0149} & 0.9694 ± 0.0042 & \textbf{0.9942 ± 0.0009} & 3.333 ± 0.124 & 1.553 ± 0.086 & 3.429 ± 0.149 \\
    & \cellcolor{lightgray}{softDice~\cite{milletari2016v} + Graph Decoder} & \cellcolor{lightgray}{0.6289 ± 0.0175} & \cellcolor{lightgray}{\textbf{0.9731 ± 0.0045}} & \cellcolor{lightgray}{0.9941 ± 0.0009} & \cellcolor{lightgray}{1.933 ± 0.065} & \cellcolor{lightgray}{1.729 ± 0.088} & \cellcolor{lightgray}{2.229 ± 0.105} \\
    & \cellcolor{lightgray}{softDice~\cite{milletari2016v} + Graph Decoder + Morph Module} & \cellcolor{lightgray}{0.6388 ± 0.0168} & \cellcolor{lightgray}{/} & \cellcolor{lightgray}{\textbf{0.9942 ± 0.0009}} & \cellcolor{lightgray}{\textbf{0.620 ± 0.021}} & \cellcolor{lightgray}{\textbf{1.355 ± 0.083}} & \cellcolor{lightgray}{\textbf{1.122 ± 0.075}} \\
    \bottomrule
  \end{tabular}
}
\end{table}

\subsection{Main Results}
We first verify the effectiveness of GraphMorph on the centerline extraction task. Then, considerable experiments are conducted on the more common segmentation task, demonstrating the powerful topological modelling capability of GraphMorph. 

\myparagraph{Centerline Extraction.} In our experiments with UNet and softDice loss on four public datasets, detailed in Table~\ref{tab:cline_results}, the inclusion of the Graph Decoder during training enables the network to learn branch-level features, leading to enhanced performance in both volumetric and topological metrics. During inference, the utilization of Morph Module results in a slight decrease in volumetric metrics; however, there is a notable enhancement in topological metrics, confirming that our network has effectively captured branch-level features of tubular structures. Overall, the combined use of the Graph Decoder and Morph Module showcases the ability to refine the segmentation network's performance, particularly in preserving the crucial topological characteristics. Our methods also beat previous SOTA Pointscatter~\cite{wang2022pointscatter} by a large margin..

\begin{table}[htbp]
  \caption{Segmentation performance based on different segmentation networks.}
  \label{tab:seg_results}
  \centering
  \renewcommand{\arraystretch}{1.0} 
  \resizebox{\textwidth}{!}{%
  \begin{tabular}{cccccccccccc}
    \toprule
    \multirow{2}{*}{Dataset} & \multirow{2}{*}{Backbone} & \multirow{2}{*}{Method} & \multicolumn{3}{c}{Volumetric metrics (\(\uparrow\))} & \multicolumn{2}{c}{Distribution metrics} & \multicolumn{3}{c}{Topological metrics (\(\downarrow\))} \\
    \cmidrule(lr){4-6} \cmidrule(lr){7-8} \cmidrule(lr){9-11}
            &          &      & Dice       & clDice     & ACC       & ARI(\(\uparrow\))       & VOI(\(\downarrow\))       & \(\beta_0\) error & \(\beta_1\) error & \(\chi\) error \\
    \midrule
    & \multirow{2}{*}{UNet~\cite{ronneberger2015u}}    & softDice~\cite{milletari2016v}      & 0.8148 ± 0.0093 & 0.8128 ± 0.0169 & 0.9535 ± 0.0023 & 0.767 ± 0.011 & 0.348 ± 0.012 & 1.191 ± 0.069 & 1.078 ± 0.074 & 1.467 ± 0.083 \\
    &         & \textbf{softDice+Ours} & \textbf{0.8238 ± 0.0091} & \textbf{0.8278 ± 0.0166} & \textbf{0.9557 ± 0.0023} & \textbf{0.778 ± 0.011} & \textbf{0.336 ± 0.012} & \textbf{0.692 ± 0.047} & \textbf{0.932 ± 0.068} & \textbf{0.951 ± 0.062} \\
    \cmidrule(lr){2-11}
    \multirow{2}{*}{DRIVE} 
    & \multirow{2}{*}{ResUNet~\cite{zhang2018road}} & softDice~\cite{milletari2016v}      & 0.8183 ± 0.0094 & 0.8183 ± 0.0178 & 0.9543 ± 0.0022 & 0.771 ± 0.011 & 0.342 ± 0.011 & 1.110 ± 0.066 & 1.059 ± 0.073 & 1.379 ± 0.079 \\
    &         & \textbf{softDice+Ours} & \textbf{0.8233 ± 0.0095} & \textbf{0.8273 ± 0.0172} & \textbf{0.9555 ± 0.0022} & \textbf{0.777 ± 0.011} & \textbf{0.336 ± 0.011} & \textbf{0.723 ± 0.047} & \textbf{0.986 ± 0.071} & \textbf{0.993 ± 0.063} \\
    \cmidrule(lr){2-11}
    & \multirow{2}{*}{CS-Net~\cite{mou2019cs}}   & softDice~\cite{milletari2016v}      & 0.8089 ± 0.0123 & 0.8073 ± 0.0185 & 0.9527 ± 0.0027 & 0.761 ± 0.014 & 0.350 ± 0.011 & 1.211 ± 0.072 & 1.096 ± 0.076 & 1.491 ± 0.085 \\
    &         & \textbf{softDice+Ours} & \textbf{0.8223 ± 0.0088} & \textbf{0.8253 ± 0.0171} & \textbf{0.9554 ± 0.0021} & \textbf{0.776 ± 0.010} & \textbf{0.336 ± 0.011} & \textbf{0.680 ± 0.044} & \textbf{0.990 ± 0.070} & \textbf{0.929 ± 0.059} \\
    \midrule 
    & \multirow{2}{*}{UNet~\cite{ronneberger2015u}}    & softDice~\cite{milletari2016v}      & 0.8043 ± 0.0092 & 0.9295 ± 0.0078 & 0.9146 ± 0.0060 & 0.653 ± 0.018 & 0.785 ± 0.040 & 0.569 ± 0.046 & 0.616 ± 0.047 & 0.738 ± 0.052 \\
    &         & \textbf{softDice+Ours} & \textbf{0.8216 ± 0.0091} & \textbf{0.9449 ± 0.0069} & \textbf{0.9211 ± 0.0057} & \textbf{0.678 ± 0.018} & \textbf{0.745 ± 0.039} & \textbf{0.361 ± 0.034} & \textbf{0.520 ± 0.043} & \textbf{0.488 ± 0.041} \\
    \cmidrule(lr){2-11}
    \multirow{2}{*}{ISBI12}
    & \multirow{2}{*}{ResUNet~\cite{zhang2018road}} & softDice~\cite{milletari2016v}      & 0.8061 ± 0.0093 & 0.9307 ± 0.0086 & 0.9153 ± 0.0055 & 0.655 ± 0.017 & 0.781 ± 0.037 & 0.572 ± 0.045 & 0.588 ± 0.048 & 0.710 ± 0.050 \\
    &         & \textbf{softDice+Ours} & \textbf{0.8166 ± 0.0105} & \textbf{0.9405 ± 0.0087} & \textbf{0.9194 ± 0.0055} & \textbf{0.671 ± 0.018} & \textbf{0.756 ± 0.037} & \textbf{0.395 ± 0.037} & \textbf{0.576 ± 0.046} & \textbf{0.518 ± 0.043} \\
    \cmidrule(lr){2-11}
    & \multirow{2}{*}{CS-Net~\cite{mou2019cs}}   & softDice~\cite{milletari2016v}      & 0.8163 ± 0.0118 & 0.9391 ± 0.0092 & 0.9194 ± 0.0075 & 0.671 ± 0.023 & 0.754 ± 0.049 & 0.451 ± 0.040 & 0.596 ± 0.046 & 0.622 ± 0.047 \\
    &         & \textbf{softDice+Ours} & \textbf{0.8282 ± 0.0118} & \textbf{0.9469 ± 0.0081} & \textbf{0.9243 ± 0.0068} & \textbf{0.690 ± 0.022} & \textbf{0.722 ± 0.045} & \textbf{0.342 ± 0.034} & \textbf{0.501 ± 0.043} & \textbf{0.452 ± 0.040} \\
    \midrule
    & \multirow{2}{*}{UNet~\cite{ronneberger2015u}}    & softDice~\cite{milletari2016v}      & 0.8170 ± 0.0402 & 0.8526 ± 0.0306 & 0.9749 ± 0.0044 & 0.781 ± 0.042 & 0.276 ± 0.033 & 0.786 ± 0.064 & 0.653 ± 0.072 & 0.960 ± 0.079 \\
    &         & \textbf{softDice+Ours} & \textbf{0.8210 ± 0.0464} & \textbf{0.8578 ± 0.0372} & \textbf{0.9756 ± 0.0045} & \textbf{0.786 ± 0.049} & \textbf{0.271 ± 0.033} & \textbf{0.545 ± 0.046} & \textbf{0.618 ± 0.067} & \textbf{0.691 ± 0.059} \\
    \cmidrule(lr){2-11}
    \multirow{2}{*}{STARE} 
    & \multirow{2}{*}{ResUNet~\cite{zhang2018road}} & softDice~\cite{milletari2016v}      & 0.7982 ± 0.0628 & 0.8343 ± 0.0512 & 0.9735 ± 0.0055 & 0.761 ± 0.065 & 0.282 ± 0.036 & 0.770 ± 0.067 & 0.707 ± 0.073 & 0.915 ± 0.079 \\
    &         & \textbf{softDice+Ours} & \textbf{0.8151 ± 0.0513} & \textbf{0.8522 ± 0.0404} & \textbf{0.9752 ± 0.0047} & \textbf{0.779 ± 0.053} & \textbf{0.273 ± 0.034} & \textbf{0.582 ± 0.054} & \textbf{0.623 ± 0.068} & \textbf{0.743 ± 0.065} \\
    \cmidrule(lr){2-11}
    & \multirow{2}{*}{CS-Net~\cite{mou2019cs}}   & softDice~\cite{milletari2016v}      & 0.7785 ± 0.0615 & 0.8173 ± 0.0474 & 0.9715 ± 0.0056 & 0.739 ± 0.063 & 0.294 ± 0.037 & 0.871 ± 0.071 & 0.803 ± 0.081 & 1.049 ± 0.087 \\
    &         & \textbf{softDice+Ours} & \textbf{0.7968 ± 0.0612} & \textbf{0.8351 ± 0.0483} & \textbf{0.9733 ± 0.0059} & \textbf{0.760 ± 0.064} & \textbf{0.282 ± 0.039} & \textbf{0.579 ± 0.049} & \textbf{0.685 ± 0.071} & \textbf{0.724 ± 0.062} \\
    \midrule
    & \multirow{2}{*}{UNet~\cite{ronneberger2015u}}    & softDice~\cite{milletari2016v}      & 0.7808 ± 0.0146 & 0.8768 ± 0.0159 & 0.9780 ± 0.0036 & 0.750 ± 0.017 & 0.239 ± 0.033 & 0.479 ± 0.020 & 0.798 ± 0.076 & 0.777 ± 0.072 \\
    &         & \textbf{softDice+Ours} & \textbf{0.7849 ± 0.0139} & \textbf{0.8816 ± 0.0151} & \textbf{0.9783 ± 0.0035} & \textbf{0.754 ± 0.016} & \textbf{0.237 ± 0.033} & \textbf{0.386 ± 0.016} & \textbf{0.754 ± 0.076} & \textbf{0.672 ± 0.070} \\
    \cmidrule(lr){2-11}
    \multirow{2}{*}{MassRoad} 
    & \multirow{2}{*}{ResUNet~\cite{zhang2018road}} & softDice~\cite{milletari2016v}      & 0.7730 ± 0.0152 & 0.8663 ± 0.0162 & 0.9773 ± 0.0036 & 0.742 ± 0.017 & 0.245 ± 0.033 & 0.799 ± 0.030 & 0.902 ± 0.078 & 1.089 ± 0.076 \\
    &         & \textbf{softDice+Ours} & \textbf{0.7755 ± 0.0150} & \textbf{0.8707 ± 0.0162} & \textbf{0.9775 ± 0.0036} & \textbf{0.744 ± 0.017} & \textbf{0.243 ± 0.033} & \textbf{0.587 ± 0.023} & \textbf{0.869 ± 0.078} & \textbf{0.869 ± 0.073} \\
    \cmidrule(lr){2-11}
    & \multirow{2}{*}{CS-Net~\cite{mou2019cs}}   & softDice~\cite{milletari2016v}      & 0.7770 ± 0.0147 & 0.8716 ± 0.0163 & 0.9779 ± 0.0035 & 0.746 ± 0.016 & 0.240 ± 0.032 & 0.487 ± 0.020 & 0.796 ± 0.075 & 0.784 ± 0.072 \\
    &         & \textbf{softDice+Ours} & \textbf{0.7789 ± 0.0149} & \textbf{0.8756 ± 0.0163} & \textbf{0.9779 ± 0.0035} & \textbf{0.748 ± 0.017} & \textbf{0.240 ± 0.033} & \textbf{0.399 ± 0.017} & \textbf{0.772 ± 0.076} & \textbf{0.682 ± 0.070} \\
    \bottomrule
  \end{tabular}
}
\end{table}

\begin{table}[htbp]
  \caption{Comparison with SOTA methods on the segmentation task. Best results are in bold; second-best are underlined. Our approach secures all leading scores and most secondary peaks.}
  \label{tab:sota_results}
  \centering
  \scriptsize
  \renewcommand{\arraystretch}{1.0}
  \resizebox{\textwidth}{!}{%
    \begin{tabular}{cccccccccccc}
      \toprule
      \multirow{2}{*}{Dataset} & \multirow{2}{*}{Backbone} & \multirow{2}{*}{Method} & \multicolumn{3}{c}{Volumetric metrics (\(\uparrow\))} & \multicolumn{2}{c}{Distribution metrics} & \multicolumn{3}{c}{Topological metrics (\(\downarrow\))} \\
      \cmidrule(lr){4-6} \cmidrule(lr){7-8} \cmidrule(lr){9-11}
      &          &      & Dice       & clDice     & ACC       & ARI(\(\uparrow\))       & VOI(\(\downarrow\))       & \(\beta_0\) error & \(\beta_1\) error & \(\chi\) error \\
      \midrule
      \multirow{8}{*}{DRIVE} 
        & UNet & softDice~\cite{milletari2016v} & 0.8148 ± 0.0093 & 0.8128 ± 0.0169 & 0.9535 ± 0.0023 & 0.767 ± 0.011 & 0.348 ± 0.012 & 1.191 ± 0.069 & 1.078 ± 0.074 & 1.467 ± 0.083 \\
        & UNet & clDice~\cite{shit2021cldice} & 0.8150 ± 0.0078 & \underline{0.8322 ± 0.0163} & 0.9520 ± 0.0021 & 0.765 ± 0.009 & 0.357 ± 0.012 & 0.910 ± 0.056 & 0.998 ± 0.070 & 1.181 ± 0.071 \\
        & UNet & Pointscatter~\cite{wang2022pointscatter} & 0.8155 ± 0.0081 & 0.8277 ± 0.0179 & 0.9525 ± 0.0020 & 0.766 ± 0.009 & 0.353 ± 0.010 & 1.360 ± 0.080 & 1.276 ± 0.083 & 1.663 ± 0.094 \\
        & UNet	& TopoLoss~\cite{hu2019topology} & \underline{0.8187 ± 0.0075} & 0.8194 ± 0.0160 & \underline{0.9540 ± 0.0020} & \underline{0.771 ± 0.009} & 0.345 ± 0.010 & 0.821 ± 0.050 & 0.997 ± 0.072 & 1.100 ± 0.067 \\
        & DSCNet~\cite{qi2023dynamic} & softDice & 0.8118 ± 0.0083 & 0.8107 ± 0.0172 & 0.9527 ± 0.0021 & 0.763 ± 0.010 & 0.352 ± 0.011 & 1.267 ± 0.075 & 1.110 ± 0.076 & 1.550 ± 0.087 \\
        & TransUNet~\cite{chen2021transunet} & softDice & 0.8153 ± 0.0094 & 0.8139 ± 0.0182 & 0.9538 ± 0.0020 & 0.768 ± 0.010 & \underline{0.344 ± 0.009} & 1.125 ± 0.066 & 1.184 ± 0.082 & 1.420 ± 0.082 \\
        & DC-UNet~\cite{lou2021dc} & softDice & 0.8086 ± 0.0103 & 0.8018 ± 0.0163 & 0.9526 ± 0.0024 & 0.760 ± 0.012 & 0.351 ± 0.011 & 1.227 ± 0.074 & 1.061 ± 0.074 & 1.499 ± 0.087 \\
        \rowcolor{lightgray}\cellcolor{white}{}
        & UNet & softDice+Ours & \textbf{0.8238 ± 0.0091} & 0.8278 ± 0.0166 & \textbf{0.9557 ± 0.0023} & \textbf{0.778 ± 0.011} & \textbf{0.336 ± 0.012} & \underline{0.692 ± 0.047} & \underline{0.932 ± 0.068} & \underline{0.951 ± 0.062} \\
        \rowcolor{lightgray}\cellcolor{white}{}
        & UNet & clDice+Ours & 0.8168 ± 0.0076 & \textbf{0.8467 ± 0.0146} & 0.9520 ± 0.0021 & 0.767 ± 0.009 & 0.357 ± 0.012 & \textbf{0.619 ± 0.043} & \textbf{0.924 ± 0.065} & \textbf{0.857 ± 0.056} \\
    \midrule
    \multirow{8}{*}{ISBI12}
      & UNet & softDice~\cite{milletari2016v} & 0.8043 ± 0.0092 & 0.9295 ± 0.0078 & 0.9146 ± 0.0060 & 0.653 ± 0.018 & 0.785 ± 0.040 & 0.569 ± 0.046 & 0.616 ± 0.047 & 0.738 ± 0.052 \\
      & UNet & clDice~\cite{shit2021cldice} & 0.8103 ± 0.0099 & 0.9353 ± 0.0084 & 0.9163 ± 0.0064 & 0.660 ± 0.020 & 0.775 ± 0.042 & 0.422 ± 0.038 & 0.563 ± 0.045 & 0.576 ± 0.043 \\
      & UNet & Pointscatter~\cite{wang2022pointscatter} & 0.8192 ± 0.0101 & 0.9406 ± 0.0077 & 0.9189 ± 0.0063 & 0.672 ± 0.020 & 0.758 ± 0.042 & 0.456 ± 0.041 & 0.568 ± 0.046 & 0.587 ± 0.047 \\
      & UNet	& TopoLoss~\cite{hu2019topology} & 0.8104 ± 0.0090 & 0.9324 ± 0.0074 & 0.9167 ± 0.0058 & 0.661 ± 0.017 & 0.773 ± 0.039 & 0.516 ± 0.041 & 0.642 ± 0.052 & 0.669 ± 0.049 \\
      & DSCNet~\cite{qi2023dynamic} & softDice & 0.8152 ± 0.0087 & 0.9366 ± 0.0078 & 0.9191 ± 0.0054 & 0.669 ± 0.016 & 0.757 ± 0.037 & 0.450 ± 0.040 & 0.567 ± 0.045 & 0.581 ± 0.044 \\
      & TransUNet~\cite{chen2021transunet} & softDice & 0.8056 ± 0.0080 & 0.9289 ± 0.0075 & 0.9148 ± 0.0055 & 0.654 ± 0.016 & 0.784 ± 0.037 & 0.636 ± 0.049 & 0.576 ± 0.047 & 0.757 ± 0.053 \\
      & DC-UNet~\cite{lou2021dc} & softDice & 0.8150 ± 0.0089 & 0.9366 ± 0.0084 & 0.9196 ± 0.0063 & 0.671 ± 0.019 & 0.753 ± 0.043 & 0.511 ± 0.043 & 0.586 ± 0.046 & 0.652 ± 0.047 \\
      \rowcolor{lightgray}\cellcolor{white}{}
      & UNet & softDice+Ours & \underline{0.8216 ± 0.0091} & \underline{0.9449 ± 0.0069} & \underline{0.9211 ± 0.0057} & \underline{0.678 ± 0.018} & \underline{0.745 ± 0.039} & \underline{0.361 ± 0.034} & \textbf{0.520 ± 0.043} & \underline{0.488 ± 0.041} \\
      \rowcolor{lightgray}\cellcolor{white}{}
      & UNet & clDice+Ours & \textbf{0.8223 ± 0.0086} & \textbf{0.9459 ± 0.0066} & \textbf{0.9213 ± 0.0056} & \textbf{0.679 ± 0.017} & \textbf{0.744 ± 0.038} & \textbf{0.353 ± 0.034} & \underline{0.539 ± 0.043} & \textbf{0.482 ± 0.040} \\
    \midrule
    \multirow{8}{*}{STARE}
      & UNet & softDice~\cite{milletari2016v} & 0.8170 ± 0.0402 & 0.8526 ± 0.0306 & 0.9749 ± 0.0044 & 0.781 ± 0.042 & 0.276 ± 0.033 & 0.786 ± 0.064 & 0.653 ± 0.072 & 0.960 ± 0.079 \\
      & UNet & clDice~\cite{shit2021cldice} & 0.8212 ± 0.0386 & 0.8579 ± 0.0319 & 0.9752 ± 0.0041 & 0.785 ± 0.040 & 0.276 ± 0.032 & 0.571 ± 0.049 & 0.629 ± 0.069 & 0.743 ± 0.065 \\
      & UNet & Pointscatter~\cite{wang2022pointscatter} & 0.8171 ± 0.0395 & 0.8533 ± 0.0331 & 0.9743 ± 0.0041 & 0.780 ± 0.041 & 0.285 ± 0.031 & 0.844 ± 0.070 & 0.781 ± 0.080 & 0.997 ± 0.086 \\
      & UNet & TopoLoss~\cite{hu2019topology} & 0.8175 ± 0.0449 & 0.8506 ± 0.0339 & 0.9750 ± 0.0045 & 0.781 ± 0.047 & 0.276 ± 0.033 & 0.659 ± 0.056 & \underline{0.615 ± 0.068} & 0.806 ± 0.069 \\
      & DSCNet~\cite{qi2023dynamic} & softDice & 0.7988 ± 0.0420 & 0.8341 ± 0.0348 & 0.9723 ± 0.0052 & 0.759 ± 0.045 & 0.296 ± 0.037 & 0.823 ± 0.068 & 0.707 ± 0.072 & 0.988 ± 0.080 \\
      & TransUNet~\cite{chen2021transunet} & softDice & 0.8046 ± 0.0474 & 0.8428 ± 0.0370 & 0.9737 ± 0.0047 & 0.767 ± 0.049 & 0.284 ± 0.034 & 0.728 ± 0.061 & 0.723 ± 0.076 & 0.884 ± 0.078 \\
      & DC-UNet~\cite{lou2021dc} & softDice & 0.7936 ± 0.0547 & 0.8300 ± 0.0426 & 0.9728 ± 0.0052 & 0.755 ± 0.057 & 0.288 ± 0.034 & 0.834 ± 0.071 & 0.721 ± 0.075 & 0.975 ± 0.082 \\
      \rowcolor{lightgray}\cellcolor{white}{}
      & UNet & softDice+Ours & \underline{0.8210 ± 0.0464} & \underline{0.8578 ± 0.0372} & \underline{0.9756 ± 0.0045} & \underline{0.786 ± 0.049} & \textbf{0.271 ± 0.033} & \underline{0.545 ± 0.046} & 0.618 ± 0.067 & \underline{0.691 ± 0.059} \\
      \rowcolor{lightgray}\cellcolor{white}{}
      & UNet & clDice+Ours & \textbf{0.8283 ± 0.0371} & \textbf{0.8747 ± 0.0284} & \textbf{0.9757 ± 0.0040} & \textbf{0.792 ± 0.039} & \underline{0.274 ± 0.032} & \textbf{0.450 ± 0.042} & \textbf{0.582 ± 0.065} & \textbf{0.598 ± 0.055} \\
    \midrule
    \multirow{8}{*}{MassRoad}
      & UNet & softDice~\cite{milletari2016v} & 0.7808 ± 0.0146 & 0.8768 ± 0.0159 & 0.9780 ± 0.0036 & 0.750 ± 0.017 & 0.239 ± 0.033 & 0.479 ± 0.020 & 0.798 ± 0.076 & 0.777 ± 0.072 \\
      & UNet & clDice~\cite{shit2021cldice} & 0.7788 ± 0.0143 & 0.8773 ± 0.0156 & 0.9775 ± 0.0037 & 0.747 ± 0.016 & 0.244 ± 0.033 & 0.512 ± 0.022 & 0.964 ± 0.090 & 0.962 ± 0.086 \\
      & UNet & Pointscatter~\cite{wang2022pointscatter} & 0.7787 ± 0.0142 & 0.8750 ± 0.0156 & 0.9778 ± 0.0035 & 0.748 ± 0.016 & 0.242 ± 0.033 & 0.620 ± 0.027 & 0.800 ± 0.076 & 0.908 ± 0.074 \\
      & UNet & TopoLoss~\cite{hu2019topology} & 0.7797 ± 0.0150 & 0.8758 ± 0.0164 & \underline{0.9781 ± 0.0035} & 0.749 ± 0.017 & \underline{0.238 ± 0.032} & 0.439 ± 0.018 & 0.780 ± 0.076 & \underline{0.727 ± 0.071} \\
      & TransUNet~\cite{chen2021transunet} & softDice & 0.7620 ± 0.0169 & 0.8588 ± 0.0182 & 0.9766 ± 0.0038 & 0.730 ± 0.019 & 0.248 ± 0.034 & 0.734 ± 0.027 & 0.933 ± 0.079 & 1.017 ± 0.075 \\
      & LinkNet34~\cite{chaurasia2017linknet} & softDice & 0.7752 ± 0.0151 & 0.8747 ± 0.0161 & 0.9775 ± 0.0036 & 0.744 ± 0.017 & 0.243 ± 0.033 & 0.489 ± 0.021 & 0.773 ± 0.076 & 0.771 ± 0.072 \\
      & D-Linknet34~\cite{zhou2018d} & softDice & 0.7752 ± 0.0149 & 0.8743 ± 0.0161 & 0.9775 ± 0.0036 & 0.744 ± 0.017 & 0.244 ± 0.033 & 0.504 ± 0.022 & \underline{0.765 ± 0.075} & 0.777 ± 0.072 \\
      \rowcolor{lightgray}\cellcolor{white}{}
      & UNet & softDice+Ours & \textbf{0.7849 ± 0.0139} & \underline{0.8816 ± 0.0151} & \textbf{0.9783 ± 0.0035} & \textbf{0.754 ± 0.016} & \textbf{0.237 ± 0.033} & \textbf{0.386 ± 0.016} & \textbf{0.754 ± 0.076} & \textbf{0.672 ± 0.070} \\
      \rowcolor{lightgray}\cellcolor{white}{}
      & UNet & clDice+Ours & \underline{0.7851 ± 0.0137} & \textbf{0.8844 ± 0.0148} & 0.9779 ± 0.0036 & \underline{0.754 ± 0.016} & 0.241 ± 0.033 & \underline{0.393 ± 0.018} & 0.879 ± 0.085 & 0.784 ± 0.082 \\
    \bottomrule
    \end{tabular}%
  }
\end{table}

\myparagraph{Segmentation.} In the initial phase of our segmentation experiments, we assessed the effectiveness of our method across different segmentation networks and datasets. As detailed in Table~\ref{tab:seg_results}, enhancements were evident in various metrics for all dataset and network combinations. The improvements in volumetric and distribution metrics underscore our method's effectiveness in the precision and reliability in segmenting and clustering accuracy. Most notably, the substantial advancements in topological metrics, particularly in the \(\beta_0\) error, highlight our method's proficiency in capturing the intricate branch-level features of tubular structures.

Moreover, the comprehensive comparisons presented in Table~\ref{tab:sota_results} across various datasets underscore the superiority of our method over current state-of-the-art techniques. The results unequivocally illustrate that our approach excels in all evaluated metrics, outstripping other methodologies. The crux of this advancement lies in the exploitation of branch-level features. Compared to our approach, traditional pixel-level classification strategies, whether employing grid representations like softDice~\cite{milletari2016v} or clDice~\cite{shit2021cldice}, or point representations akin to Pointscatter~\cite{wang2022pointscatter}, inherently lack in capturing the essential branch-level features. Innovatively, Our method incorporates the graph representation to capture explicit branch-level features, and morphs the predicted graphs to topologically accurate centerline masks, which are subsequently utilized for post-processing. The excellent performance of various metrics, especially topological metrics, proves the validity of our approach.

\subsection{Ablation Study}
\myparagraph{Size of ROI (H).} See Table~\ref{tab:roi_effect}, the experimental results on the DRIVE dataset suggest that a default ROI size of \( H=32 \) provides a favorable balance across the evaluation metrics. Specifically, when \(H\) is reduced, an increase in \(\beta_0\) error is observed, indicating diminished topological accuracy. Conversely, increasing \(H\) does not offer substantial metric improvements and leads to higher computational demands. Hence, \( H=32 \) is adopted as the default ROI size for all three medical datasets. Similarly, the default ROI size for the MassRoad dataset has been determined to be \( H=48 \).

\myparagraph{Threshold in the Morph Module (\(p_{thresh}\)).} We conduct experiments on the centerline extraction task. A smaller value of \( p_{thresh} \) denotes a more stringent selection criterion, with \( p_{thresh} = 1.0 \) indicating that all paths are considered without any selection filter. As detailed in Table~\ref{tab:p_thresh_effect}, the best results were achieved at \( p_{thresh} = 0.5 \), which is adopted as the default setting across all experiments.

\myparagraph{Post-processing on the Segmentation Task.} 
As shown in Table~\ref{tab:postproc_effect}, incorporating post-processing has led to a slight improvement in volumetric metrics and a significant elevation in topological metrics. Notably, the substantial enhancement in the \(\beta_0\) metric primarily results from the successful suppression of false positives, which aligns with our initial hypothesis. For a qualitative demonstration, we direct the reader to the visual comparisons presented in the Appendix~\ref{appendix_vis_postprocess}.

\begin{figure}[htbp]
  \centering
  \begin{minipage}[t]{0.48\textwidth}
  \captionof{table}{Effect of ROI size \(H\) on two tasks.}
  \label{tab:roi_effect}
  \renewcommand{\arraystretch}{1.22}
  \resizebox{\textwidth}{!}{%
  \begin{tabular}{ccccccc}
    \toprule
    \multirow{2}{*}{Dataset} & \multirow{2}{*}{\(H\)} & \multicolumn{3}{c}{Segmentation} & \multicolumn{2}{c}{Centerline Extraction} \\
    \cmidrule{3-5}\cmidrule{6-7}
     & & Dice & clDice & \(\beta_0\) error & Dice & \(\beta_0\) error \\
    \midrule
    \multirow{4}{*}{DRIVE} & 16 & 82.61 & 82.96 & 0.763 & 74.36 & 0.771 \\
     & \cellcolor{lightgray}{32} & \cellcolor{lightgray}{82.44} & \cellcolor{lightgray}{82.78} & \cellcolor{lightgray}{0.692} & \cellcolor{lightgray}{74.97} & \cellcolor{lightgray}{0.555} \\
     & 48 & 82.43 & 82.78 & 0.676 & 74.90 & 0.492 \\
     & 64 & 82.44 & 82.63 & 0.666 & 74.72 & 0.540 \\
    \midrule
    \multirow{4}{*}{MassRoad} & 16 & 78.16 & 87.66 & 0.457 & 61.36 & 2.265 \\
     & 32 & 78.33 & 87.90 & 0.414 & 62.90 & 0.944 \\
     & \cellcolor{lightgray}{48} & \cellcolor{lightgray}{78.53} & \cellcolor{lightgray}{88.16} & \cellcolor{lightgray}{0.386} & \cellcolor{lightgray}{63.59} & \cellcolor{lightgray}{0.620} \\
     & 64 & 78.35 & 87.93 & 0.396 & 63.27 & 0.539 \\
    \bottomrule
    \end{tabular}%
}
  \end{minipage}
  \hfill
  \begin{minipage}[t]{0.48\textwidth}
  \captionof{table}{Effect of \(p_{thresh}\).}
  \label{tab:p_thresh_effect}
  \resizebox{\textwidth}{!}{%
  \begin{tabular}{ccccccc}
    \toprule
    Dataset & p\_thresh & Dice & ACC & \(\beta_0\) error & \(\beta_1\) error & \(\chi\) error \\
    \midrule
    \multirow{4}{*}{DRIVE} & 0.25 & 70.90 & 97.67 & 0.815 & 1.528 & 1.121 \\
    & \cellcolor{lightgray}{0.5} & \cellcolor{lightgray}{74.97} & \cellcolor{lightgray}{97.76} & \cellcolor{lightgray}{0.555} & \cellcolor{lightgray}{1.074} & \cellcolor{lightgray}{0.893} \\
    & 0.75 & 74.97 & 97.71 & 0.572 & 1.221 & 1.025 \\
    & 1.0 & 74.80 & 97.68 & 0.604 & 1.270 & 1.069 \\
    \bottomrule
  \end{tabular}
    }
    \vfill
    \captionof{table}{Effect of post-processing.}
      \label{tab:postproc_effect}
      \resizebox{\textwidth}{!}{%
      \begin{tabular}{ccccccc}
        \toprule
        Dataset & Method & Dice & clDice & $\beta_0$ error & $\beta_1$ error & $\chi$ error \\
        \midrule
        \multirow{2}{*}{DRIVE} & softDice+Ours & \textbf{82.44} & \textbf{82.78} & \textbf{0.692} & \textbf{0.932} & \textbf{0.951} \\
        & w.o. Post-processing & 82.41 & 82.68 & 0.932 & 0.932 & 1.199 \\
        \midrule
        \multirow{2}{*}{MassRoad} & softDice+Ours & 78.44 & \textbf{87.88} & \textbf{0.374} & \textbf{0.756} & \textbf{0.655} \\
        & w.o. Post-processing & \textbf{78.45} & 87.83 & 0.480 & 0.757 & 0.757 \\
        \bottomrule
      \end{tabular}%
    }
  \end{minipage}
\end{figure}

\subsection{Qualitative Results}
\begin{figure}[htbp]
  \centering
  \includegraphics[width=\textwidth]{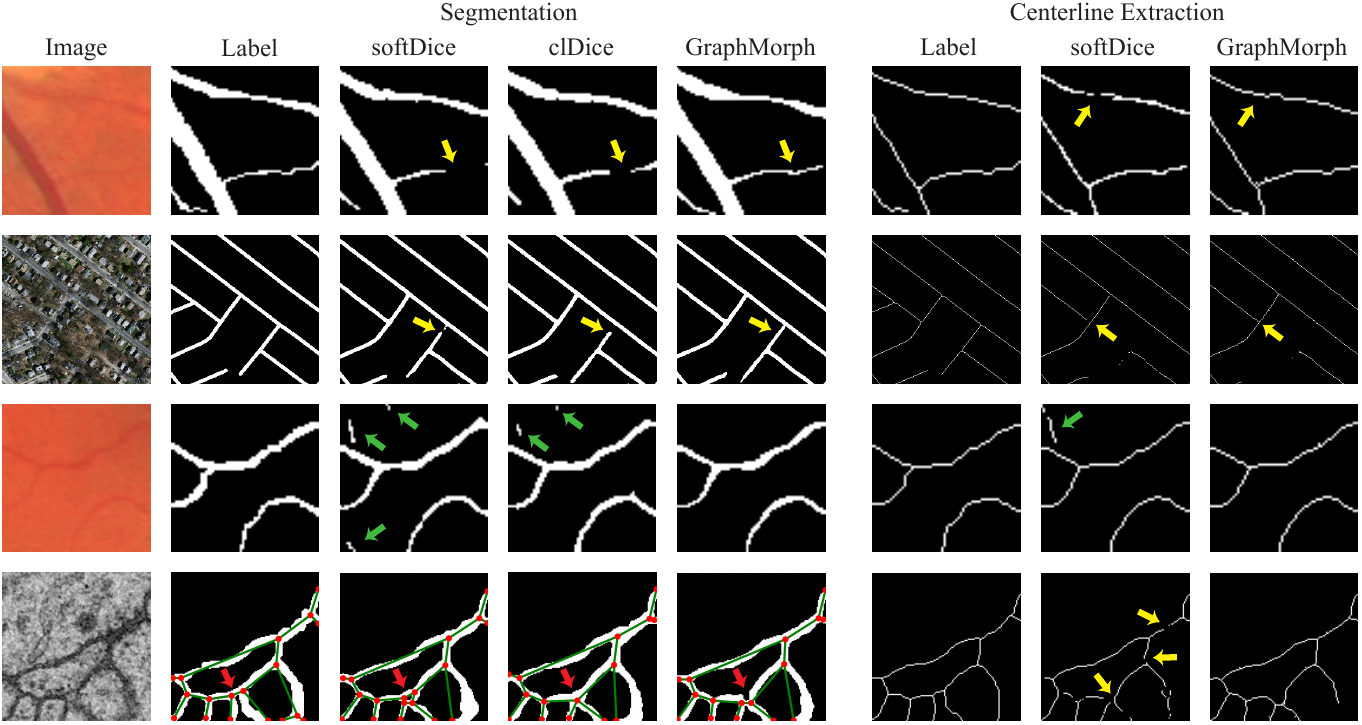}
  \caption{Visual comparison for our GraphMorph with other methods (zoom for details). Areas indicated by yellow arrows show false negatives (FNs), areas pointed by green arrows demonstrate false positives (FPs), and regions highlighted by red arrows are topological errors (TEs) identifiable in other methods but are accurately resolved by our approach.}
  \label{fig:exp_vis}
\end{figure}

We qualitatively analyze the role of GraphMorph in reducing FNs, FPs, and TEs on the segmentation and centerline extraction tasks, as detailed in Figure~\ref{fig:exp_vis}. The results show that our method can effectively reduce the three types of errors in both tasks. This is due to the utilization of branch-level features during training and the effective design of inference process. Furthermore, we visualize the graphs predicted by the Graph Decoder as well as the role of the Morph Module in suppressing various errors in Appendix~\ref{append_vis_graph}.

\section{Conclusion}
\label{sec:Conclusion}

This paper introduces GraphMorph, a framework that diverges from traditional pixel-level prediction methods in tubular structure extraction. By integrating two core components, the Graph Decoder and the Morph Module, GraphMorph adaptly captures and leverages branch-level features. Equipped with our proposed link predcition module and \(\mathrm{SkeletonDijkstra}\) algorithm, the training and inference processes of the network are efficiently carried out. For the segmentation task, it further employs a straightforward yet effective post-processing strategy that substantially reduces false positives in the predictions. Extensive evaluations across various datasets for medical image segmentation and road network extraction have demonstrated the superiority of GraphMorph over existing methods, particularly in terms of topological metrics. This breakthrough not only boosts precision in application-specific tasks but also sets a robust foundation for future research in tubular structure extraction.

\section*{Acknowledgements}

This work is supported by National Science and Technology Major Project (2022ZD0114902) and National Science Foundation of China (NSFC62276005).

\bibliographystyle{plain}
\bibliography{main}

\newpage
\appendix
\renewcommand \thepart{} 
\renewcommand \partname{}
\part{Appendix}

\section{Implementation Details}
\label{appendix_implementation}

In this section, we provide more implementation details of GraphMorph.

For fair comparison to previous works like PointScatter~\cite{wang2022pointscatter}, we use the ADAM optimizer with the initial learning rate 1e-3 and cosine learning rate schedule with warm-up strategy to train the network. The weight decay is set
to be 1e-4 uniformly. We train the network for 3K iterations for the three medical image datasets, and 10K for MassRoad. We use batchsize=4 for all datasets. We implement GraphMorph based on PyTorch~\cite{paszke2019pytorch} and Detectron2~\cite{wu2019detectron2}.

Our \(\mathrm{SkeletonDijkstra}\) algorithm is designed to run solely on CPU due to its computational nature.We use \(p_{thresh}=0.5\) across all experiments. To enhance performance and efficiency, we have implemented this algorithm in C++. For a detailed understanding of the algorithm, refer to the pseudo-code provided in Appendix~\ref{appendix_SkeletonDijkstra}.

\section{Graph Construction}
\label{ssec:graph_construction}

The process of constructing a graph can be briefly summarized in three steps as follows: (1) Generate the centerline mask of the tubular structure using skeletonization algorithm~\cite{zhang1984fast}; (2) Analyze each centerline point \(P\) by counting the centerline points among its eight neighbors (denoted as \(N\), not including \(P\)). Define \(P\) as a junction if \(N \geq 3\) and as an endpoint if \(N = 1\). Points with \(N=2\) are path points and are not considered as nodes. Junctions and endpoints form the node set \(V\) of the graph \(G\). (3) If there is a pathway consisting of only path points between two nodes, then there is an edge between them. All edges form the edge set \(E\). We use publicly accessible implementations of skeletonization\footnote{\href{https://github.com/scikit-image/scikit-image/blob/v0.22.0/skimage/morphology/_skeletonize.py}{skimage.morphology.skeletonize}} and graph construction.\footnote{\href{https://github.com/Image-Py/sknw}{https://github.com/Image-Py/sknw}}

\begin{figure}[htbp]
  \centering
  \begin{subfigure}[b]{0.87\textwidth}
      \includegraphics[width=\textwidth]{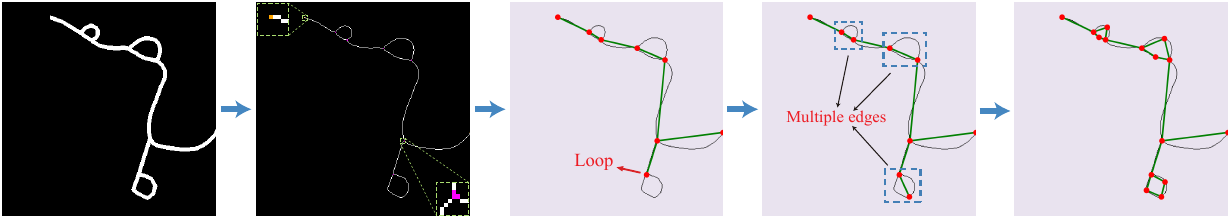}
      \caption{}
        \label{fig:a}
  \end{subfigure}
  \hfill
  \begin{subfigure}[b]{0.1\textwidth}
  \includegraphics[width=\textwidth]{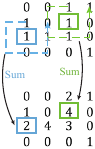}
      \caption{}
        \label{fig:b}
  \end{subfigure}
  \caption{(a) Stages of graph construction from the binary mask of a road network. The first stage demonstrates skeletonization process to a centerline mask. In the second image, we highlight the endpoints in orange and junctions in purple. Adjacent junctions are merged and considered as a single junction. Subsequent stages illustrate resolving Loops and reducing Multiple edges. (b) Example of calculating N.}
  \label{fig:build_graph}
  \vspace{-10pt}
\end{figure}

However, the resultant graph may contain elements such as \emph{loops} (closed paths where a node connects back to itself) and \emph{multiple edges} (more than one edge connecting the same pair of nodes). Addressing both these elements is crucial; otherwise, reconstructing such structures during the inference process would be challenging. As depicted in Figure \ref{fig:build_graph}, to manage these complexities, we introduce new nodes in the following manner:

1. \emph{Loops}: We insert new nodes at selected points within the loop to break the cycle.

2. \emph{Multiple edges}: After resolving loops, we then add nodes along edges where multiple connections exist between the same pair of nodes.

These modifications ensure that the graph structure is simplified and ready for more effective processing during inference.

\section{Link Prediction Module}
\label{appendix_link_prediction}

To validate the effectiveness of our dynamic link prediction module, we compare it with the approach used in RelationFormer~\cite{shit2022relationformer}.

RelationFormer learns an additional [\texttt{rln}]-token during the training of DETR to encode the relationships between node queries. In the inference stage, to predict the connection between two nodes, the method concatenates the features of the two nodes with the [\texttt{rln}]-token into a single vector. This vector is then processed by a three-layer MLP to predict the probability of connection between the nodes. Let us assume that there are \( P \) matched queries, denoted as \( \widetilde{Q}^r \in \mathbb{R}^{P \times C} \). In RelationFormer, the connection probability for each node pair is computed individually, resulting in a computational complexity of \( O(P^2 \times C^2) \). In contrast, our dynamic module achieves a complexity of \( O(P \times C^2) \) (Equation~\eqref{eq:link1} to~\eqref{eq:link4}), reducing the computational burden.

Table~\ref{tab:rln_dynamic_compare} presents a comparison between the [\texttt{rln}]-token and our dynamic module on the task of centerline extraction. The metrics "Node Detection" and "Edge Detection" are reproduced from RelationFormer,  which measure the accuracy of the nodes and edges extracted by the Graph Decoder. Other metrics assess the accuracy of the centerline masks output by the Morph Module. All experiments are based on UNet. The results demonstrate that both methods achieve comparable performance. However, our method significantly reduces the computational complexity, thereby shortening the inference time. This enhancement makes our approach more suitable for applications requiring fast processing speeds without sacrificing performance.

\begin{table}
  \caption{Comparison of [\texttt{rln}]-token and our dynamic module on the DRIVE and STARE datasets. The "Time" metric represents the cumulative time required to process all sliding windows in the link prediction phase for a single 384$\times$384 image patch during inference.}
  \label{tab:rln_dynamic_compare}
  \centering
  \resizebox{\textwidth}{!}{%
    \begin{tabular}{cccccccccccc}
    \toprule
    \multirow{2}{*}{Dataset} & \multirow{2}{*}{Method} & \multirow{2}{*}{Time/s} & \multicolumn{2}{c}{Node Detection (\(\uparrow\))} & \multicolumn{2}{c}{Edge Detection (\(\uparrow\))} & \multicolumn{2}{c}{Volumetric metrics (\(\uparrow\))} & \multicolumn{3}{c}{Topological metrics (\(\downarrow\))} \\
    \cmidrule(lr){4-5} \cmidrule(lr){6-7} \cmidrule(lr){8-9} \cmidrule(lr){10-12}
    & & & AP@0.5 & AR@0.5 & AP@0.5 & AR@0.5 & Dice & ACC & $\beta_0$ error & $\beta_1$ error & $\chi$ error \\
    \midrule
    \multirow{2}{*}{DRIVE} & [\texttt{rln}]-token & 0.2823 & 52.40 & 58.20 & 23.33 & 37.78 & 74.95 & 97.76 & 0.548 & 1.026 & 0.866 \\
                           & Dynamic   & 0.1582 & 52.30 & 58.11 & 23.25 & 37.89 & 74.97 & 97.76 & 0.555 & 1.074 & 0.893 \\
    \midrule
    \multirow{2}{*}{STARE} & [\texttt{rln}]-token & 0.1385 & 54.95 & 61.44 & 27.84 & 43.73 & 73.99 & 98.92 & 0.483 & 0.731 & 0.663 \\
                           & Dynamic   & 0.0754 & 55.06 & 61.90 & 27.80 & 43.62 & 74.25 & 98.94 & 0.482 & 0.799 & 0.653 \\
    \bottomrule
    \end{tabular}%
  }
  \vspace{-10pt}
\end{table}

\section{SkeletonDijkstra Algorithm}
\label{appendix_SkeletonDijkstra}

\begin{algorithm}[ht]
\caption{SkeletonDijkstra Algorithm}
\label{algo:skeleton_dijkstra}
\begin{algorithmic}
\REQUIRE Start point \( s \), End point \( e \), Cost map \( C \), Path threshold \( p_{thresh} \)
\ENSURE Minimum cost path from \( s \) to \( e \) under threshold \( p_{thresh} \)

\STATE Initialize priority queue \( Q \) with \( (0, [s]) \)
\STATE Initialize visited set \( Vis \)

\WHILE{not \( Q \) empty}
    \STATE \( (cost, path) \) $\leftarrow$ \( Q.pop() \)
    \STATE \( curr \) $\leftarrow$ \( \text{last element of } path \)

    \STATE Add \( curr \) to \( Vis \)

    \IF{\( curr = e \)}
        \STATE \( avg \) $\leftarrow$ \( cost / \) length(\( path \))
        \IF{\( avg > p_{thresh} \)}
            \STATE \textbf{return} \(\varnothing\)
        \ENDIF
        \STATE \textbf{return} \( path \)
    \ENDIF

    \FOR{each \( n \) in neighbors of \( curr \)}
        \IF{\( n \) in \( Vis \)}
            \STATE \textbf{continue}
        \ENDIF
        \STATE \( neis\_in\_path \) $\leftarrow$ count of \( n \)'s neighbors in \( path \)
        \IF{\( neis\_in\_path \leq 1 \)}
            \STATE \( path \) $\leftarrow$ \( path \) concatenated with \( [n] \)
            \STATE \( cost \) $\leftarrow$ \( cost + C[n.x][n.y] \)
            \STATE \( Q.push((cost, path)) \)
        \ENDIF
    \ENDFOR
\ENDWHILE

\end{algorithmic}
\end{algorithm}

The pseudo-code of our \(\mathrm{SkeletonDijkstra}\) algorithm is given in Algorithm~\ref{algo:skeleton_dijkstra}, which finds the optimal path satisfying the skeleton nature for two points.

\section{Details of Processing in Segmentation}
\label{appendix_SegProcess}

\subsection{Soft Skeletonization}
\label{appendix_SoftSke}
See \(\mathrm{soft\_skeleton}\) function in Listing~\ref{lst:ske_and_post}. In the segmentation task, the segmentation network outputs the segmentation probability \(S_m\), which we need to soft skeletonized into the centerline probability \(P_m\) for input into the Morph Module.

\subsection{Post-processing to Suppress False Postives}
\label{appendix_SuppressFPs}
See \(\mathrm{dilate\_with\_seg\_limit}\) function in Listing~\ref{lst:ske_and_post}. In the segmentation task, after obtaining topologically accurate centerline masks by the Morph Module, false positives can be greatly suppressed with this post-processing strategy.

\begin{listing}
\caption{Python codes of the soft skeletonization operation and post-processing strategy during the inference process of the segmentation task.}
\begin{lstlisting}[language=Python]
import numpy as np
import cv2
from skimage.morphology import skeletonize
from scipy.ndimage import distance_transform_edt

def soft_skeleton(seg_prob: np.ndarray):
    '''
    Generate centerline probability map from the segmentation probability map.
    seg_prob: segmentation probability map output from the segmentation network.
    '''
    seg_mask = seg_prob > 0.5
    ske = skeletonize(seg_mask)
    distmap = distance_transform_edt(~ske.astype(np.bool_))
    distmap = np.clip(distmap, 0, 64) / 64
    cline_prob = seg_prob * (1 - distmap)
    return cline_prob

def dilate_with_seg_limit(cline_mask, seg_mask, kernel_size=3):
    '''
    Post process in the segmentation task to suppress false postives.
    cline_mask: centerline mask output from the Morph module.
    seg_mask: segmentation mask thresholded from the segmentation probability map.
    kernel_size: kernel size of the dilation operation.
    '''
    kernel = np.ones((kernel_size, kernel_size), np.uint8)
    res = np.minimum(cline_mask, seg_mask)
    while True:
        dilated_cline_mask = cv2.dilate(res, kernel, iterations=1)
        dilated_res = dilated_cline_mask - res
        dilated_res = np.minimum(dilated_res, seg_mask)
        dilated_cline_mask = res + dilated_res
        if np.array_equal(dilated_cline_mask, res):
            break
        res = dilated_cline_mask
    return dilated_cline_mask
\end{lstlisting}
\label{lst:ske_and_post}
\end{listing}

\section{Computational Resources}
\label{appendix_ComputeResources}

\myparagraph{Hardware Configuration.} Experiments were conducted using an NVIDIA GeForce RTX 3090 with 24 GB GPU memory. The CPU used was an Intel(R) Xeon(R) CPU E5-2680 v4 @ 2.40GHz, which features 28 cores.

\myparagraph{Analysis of training process.} Training on the DRIVE dataset with a UNet backbone and a batch size of 4 using "SoftDice+Ours" method requires approximately 11.8 GB of GPU memory, compared to 5.4 GB of "SoftDice". More comparison between these two methods are show in Table~\ref{tab:training_comparison}. The increase in parameters and FLOPs in our approach primarily stems from the integration of the Graph Decoder featuring a DETR module. This component is crucial for predicting accurate topological structures of the graphs. Advancements in transformer architectures that reduce computational overhead could potentially enhance the efficiency of our model during training.

\myparagraph{Inference time analysis.} The inference times for each model component are summarized in Table~\ref{tab:inference_timing}. The data presented in the table was obtained by processing a 384 $\times$ 384 image patch from the DRIVE dataset. ROI size is \(H=32\), with a sliding window stride of 30. As shown in the table, significant time is concentrated on the Morph Module. The primary time expenditure currently arises from processing the patches sequentially in our sliding window strategy. However, as each patch operates independently, there is significant potential to enhance efficiency by parallelizing the computations of all patches. Recognizing this opportunity, we plan to focus future work on optimizing the Morph Module by implementing parallel processing techniques to accelerate inference.

\begin{table}[htbp]
  \caption{Comparison of required resources during training.}
  \label{tab:training_comparison}
  \centering
  \resizebox{0.6\textwidth}{!}{%
  \begin{tabular}{ccccc}
    \toprule
    Method & Params & FLOPs & Time per iteration (s) & GPU Memory \\
    \midrule
    SoftDice & 39M & 187G & 0.203 & 5.4 GB \\
    softDice+Ours & 48M & 268G & 0.589 & 11.8 GB \\
    \bottomrule
  \end{tabular}%
  }
\end{table}

\begin{table}[htbp]
  \caption{Inference timing for each Module.}
  \label{tab:inference_timing}
  \centering
  \resizebox{0.45\textwidth}{!}{%
  \begin{tabular}{ccccccc}
    \toprule
    Module & Device & Time (s) \\
    \midrule
    Segmentation network & GPU & 0.0101 \\
    Deformable DETR & GPU & 0.1194 \\
    Link prediction module & GPU & 0.0379 \\
    Morph Module & CPU & 0.2933 \\
    (Segmentation) Post-processing & CPU & 0.0279 \\
    \bottomrule
  \end{tabular}%
  }
\end{table}

\section{More Visualization Results}
\label{appendix_MoreVisualization}

\subsection{Visualization of Predicted Graphs}
\label{append_vis_graph}

\begin{figure}
  \centering
  \includegraphics[width=\textwidth]{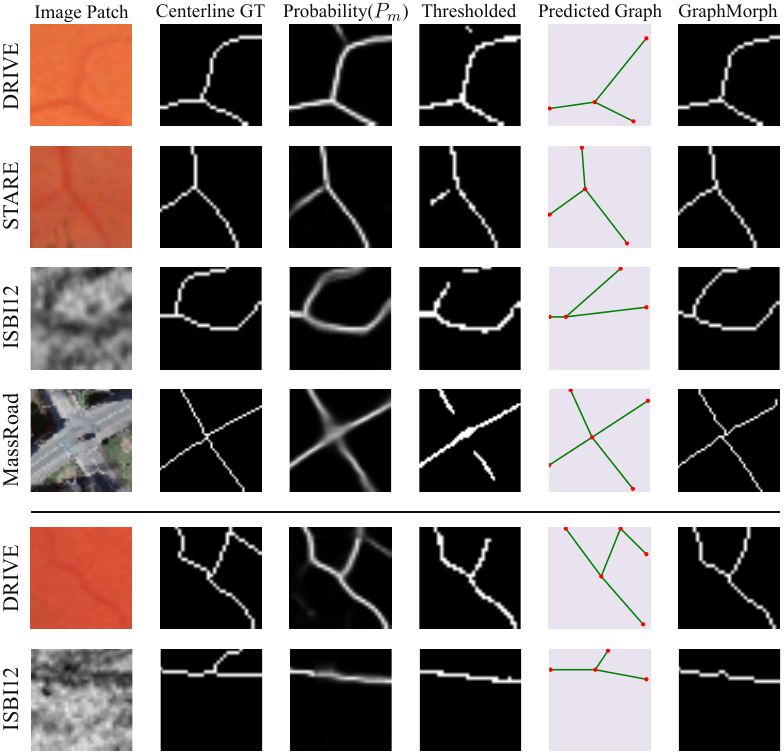}
  \caption{Visualization of intermediate results in the centerline extraction task. The results in the fourth column are obtained by thresholding \( P_m \) at 0.5. Comparisons across the first four rows illustrate that GraphMorph achieves improved results through morphing predicted graphs. The last two rows demonstrate how the settings of \( p_{\text{thresh}} \) in the Morph Module may lead to concessions to \( P_m \), resulting in false negatives.}
  \label{fig:predicted_graphs}
\end{figure}

We visualize the graphs predicted by the Graph Decoder within the context of the centerline extraction task and analyze the role of the Morph Module. As shown in Figure~\ref{fig:predicted_graphs}, the first four rows illustrate the Graph Decoder's robust capability to predict graphs. By comparing the final predicted results obtained through the Morph Module (last column) with those obtained by thresholding \(P_m\) at 0.5 (fourth column), it is evident that issues such as redundant and broken branches are effectively mitigated. However, the Morph Module also has limitations. Notably, the setting of \( p_{\text{thresh}} \) might lead to overlooking some true-positive edges due to inaccuracies in \( P_m \), which is illustrating in the last two rows in Figure~\ref{fig:predicted_graphs}. This highlights areas for future improvement.

\subsection{Visualization of Effect of Post-processing on the Segmentation Task}
\label{appendix_vis_postprocess}
\begin{figure}
  \centering
  \includegraphics[width=0.8\textwidth]{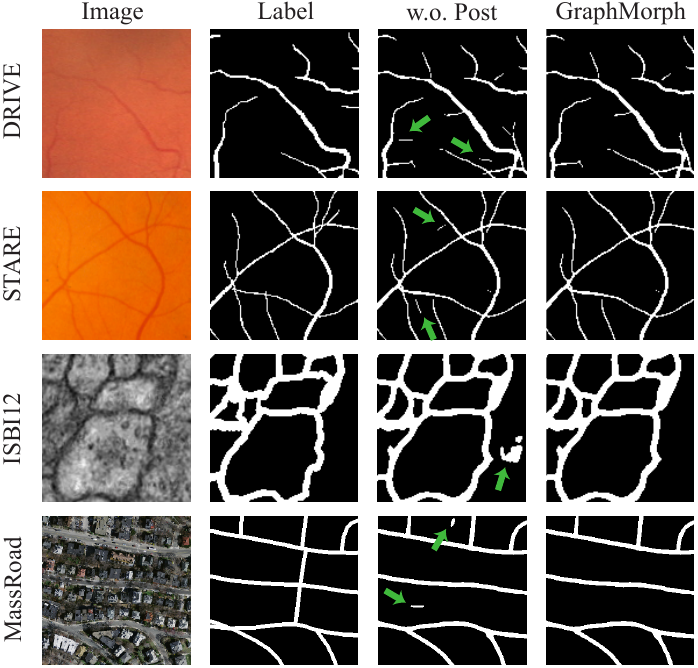}
  \caption{Visualization of the effect of the post-processing in the segmentation task across four datasets. Columns represent, from left to right: original images, ground truth segmentation labels, thresholded output from the segmentation network (w.o. Post), and results with post-processing. Green arrows highlight areas where false positives have been successfully suppressed.}
  \label{fig:appendix_post_process}
\end{figure}

Figure~\ref{fig:appendix_post_process} showcases the impact of post-processing in mitigating false positives within segmentation tasks, a procedure fully elaborated in Appendix~\ref{appendix_SuppressFPs}. This figure clearly reveals the elimination of isolated regions, originally predicted by the segmentation networks, across all datasets. Notably, the excision of such regions—often minute in scale—exerts a nominal effect on volumetric metrics while markedly bolstering topological metrics. This enhancement in the integrity of topological metrics through post-processing is substantiated by the data in Table~\ref{tab:postproc_effect}.

\section{Limitations}
\label{append_limitations}
Despite the advancements offered by GraphMorph, the method exhibits certain limitations. First, the reliance on post-processing for the segmentation task indicates a potential underutilization of branch-level features. Although false postives are significantly suppressed, segmentation results are not always topologically aligned with predicted graphs, suggesting room for improvement in segmentation performance. Additionally, the necessity to train on relatively small ROIs, due to the complex nature of tubular structures, requires sliding window technique during inference. This technique may not fully capture comprehensive branch-level details and the global context of the entire tubular structure. Based on the limitations, future developments will aim to refine segmentation algorithms to utilize predicted graphs directly, thereby reducing dependency on post-processing. Concurrently, efforts will also focus on the capability of processing larger fields of view in a single analysis, thus preserving global context and enhancing feature consistency across the entire structure.

\section{Additional Experiments on 3D Dataset}
\label{append_parse}

\begin{figure}[htbp]
  \centering
  \includegraphics[width=0.75\textwidth]{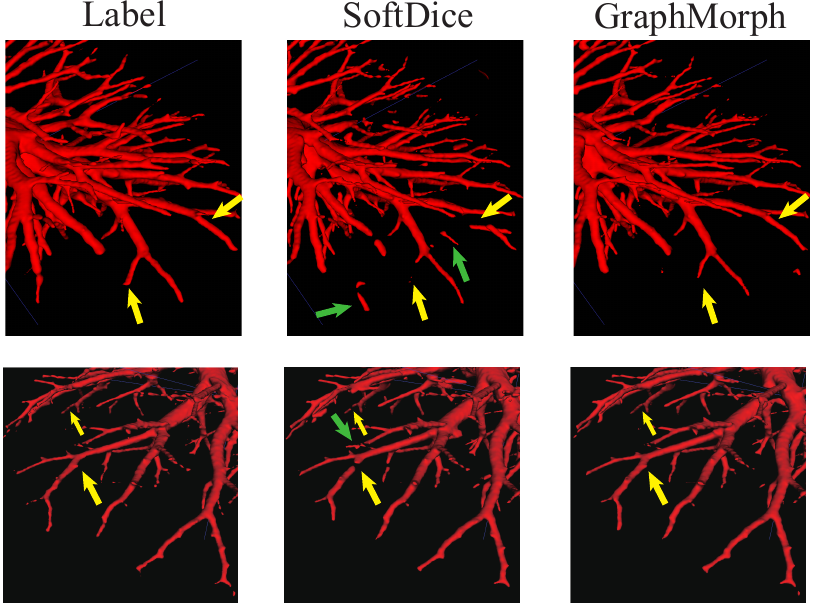}
  \caption{Visual comparison for our GraphMorph with baseline on the segmentation task. Areas indicated by yellow arrows show false negatives (FNs) and areas pointed by green arrows demonstrate false positives (FPs) appear in baseline but are accurately predicted by our approach.}
  \label{fig:append_parse_vis}
\end{figure}

The application of GraphMorph to 3D medical datasets can be initially explored for its clinical significance. Thus, we have extended GraphMorph to use the 3D UNet architecture and tested it on the pulmonary arterial vascular segmentation dataset from the PARSE challenge, which includes 100 annotated 3D CT scans. These cases were divided in a 7:1:2 ratio for training, validation, and testing.

The preliminary results, as detailed in Table~\ref{tab:append_parse_results}, show that our method consistently outperforms existing baselines across all metrics, mirroring the success we observed with 2D data. This alignment between 2D and 3D results not only underlines the effectiveness of our method but also its adaptability to 3D vessel segmentation task, which indicates the potential of GraphMorph in clinical diagnosis. Moreover, Figure~\ref{append_parse} demonstrates that GraphMorph effectively suppresses false positives (FPs) and false negatives (FNs) in fine structures. Further attempts on 3D datasets will be made to validate its effectiveness.

\begin{table}[htbp]
  \caption{Segmentation performance of GraphMorph on PARSE dataset.}
  \label{tab:append_parse_results}
  \centering
  \renewcommand{\arraystretch}{1.2} 
  \resizebox{\textwidth}{!}{%
  \begin{tabular}{cccccccccc}
    \toprule
    \multirow{2}{*}{Backbone} & \multirow{2}{*}{Method} & \multicolumn{3}{c}{Volumetric metrics (\(\uparrow\))} & \multicolumn{2}{c}{Distribution metrics} & \multicolumn{2}{c}{Topological metrics (\(\downarrow\))} \\
    \cmidrule(lr){3-5} \cmidrule(lr){6-7} \cmidrule(lr){8-9}
            &      & Dice       & clDice     & ACC       & ARI(\(\uparrow\))       & VOI(\(\downarrow\))       & \(\beta_0\) error & \(\chi\) error \\
    \midrule
    \multirow{2}{*}{UNet}    & softDice      & 0.7968 ± 0.0166 & 0.8350 ± 0.0152 & 0.9878 ± 0.0021 & 0.779 ± 0.017 & 0.134 ± 0.019 & 1.087 ± 0.078 & 1.130 ± 0.082 \\
    & \textbf{softDice+Ours} & \textbf{0.8196 ± 0.0138} & \textbf{0.8730 ± 0.0111} & \textbf{0.9901 ± 0.0015} & \textbf{0.805 ± 0.014} & \textbf{0.115 ± 0.015} & \textbf{0.536 ± 0.039} & \textbf{0.602 ± 0.045} \\
    \bottomrule
  \end{tabular}
}
\end{table}

\section{Broader Impacts}
\label{appendix_BroaderImpacts}

In this work, we present GraphMorph, a framework aimed at improving the extraction of tubular structures in medical image analysis, such as blood vessels and other elongated anatomical features. Fine-scale structures often consist of interconnected branches forming cohesive networks critical to physiological functions. By enhancing topological accuracy, GraphMorph provides more coherent and precise representations of these structures. These improved predictions with better topology in medical diagnostic scenarios related to tubular structures may assist clinical diagnosis. While our results are promising, they are based on publicly available datasets that may not fully capture the complexity and variability of real-world clinical data; therefore, further validation on more 3D datasets is necessary to confirm its applicability in clinical settings. At the present stage, we do not foresee any potential negative societal impacts arising from our work. Our goal is to contribute a useful tool for the medical imaging community, supporting efforts to improve segmentation accuracy and ultimately aiding in better healthcare outcomes.

\end{document}